\documentclass[twoside]{article}
\usepackage[utf8]{inputenc}

\usepackage{natbib}
\usepackage{booktabs}
\usepackage{appendix}

\usepackage{enumitem}
\usepackage{todonotes}

\usepackage[colorlinks = true,
             linkcolor = blue,
             urlcolor  = blue,
             citecolor = blue,
             anchorcolor = blue]{hyperref}
\usepackage{url}
\usepackage{amsmath}
\usepackage{subcaption}
\usepackage{enumitem}
\usepackage[a4paper]{geometry}
\usepackage{siunitx}

\usepackage{pythonhighlight}
\usepackage{floatrow}
\usepackage{nicefrac}       % compact symbols for 1/2, etc.
\usepackage{microtype}      % microtypography
\usepackage{amsmath}
\usepackage{amsfonts}
\usepackage{algorithmic}
\usepackage{mathtools}
\usepackage{amsthm}
\usepackage{indentfirst}
\usepackage{algorithm}
\usepackage{algorithmic}
\usepackage[normalem]{ulem}
\useunder{\uline}{\ul}{}

\usepackage{bm}

\newcommand{\cA}{\mathcal{A}}
\newcommand{\cS}{\mathcal{S}}

\newcommand{\cD}{\mathcal{D}}

\newcommand{\cU}{\mathcal{U}}
\newcommand{\cX}{\mathcal{X}}

\newcommand{\cK}{\mathcal{K}}

\newcommand{\mathbbm}[1]{\text{\usefont{U}{bbm}{m}{n}#1}}

\let\epsilon\relax
\DeclareMathOperator{\epsilon}{\varepsilon}

\usepackage{RR}

\RRdate{July 2022}

\RRauthor{
    Romain Gautron\thanks{\begin{minipage}[t]{\linewidth}AIDA, Univ Montpellier, CIRAD, Montpellier, France \& CGIAR Platform for Big Data in Agriculture, Alliance of Bioversity International and CIAT, Km 17, Recta Cali-Palmira 763537, Colombia\\ \url{romain.gautron@cirad.fr}\end{minipage}}
    \and Emilio J. Padrón\thanks{\begin{minipage}[t]{\linewidth}UDC--Computer Architecture Group \& CITIC (Center for ICT Research) \& Edif. Área Científica, Campus Elviña S/N 15071, A Coruña, Spain\\
    \url{emilio.padron@udc.gal}\end{minipage}}
    \and Philippe Preux\thanks{\begin{minipage}[t]{\linewidth}Université de Lille, CNRS, Inria, F-59650 Villeneuve d'Ascq, France\\
    \url{philippe.preux@inria.fr}\end{minipage}}
    \and Julien Bigot\thanks{\begin{minipage}[t]{\linewidth}Université Paris-Saclay, UVSQ, CNRS, CEA, Maison de la Simulation, 91191, Gif-sur-Yvette, France\\ \url{julien.bigot@cea.fr}\end{minipage}}
    \and Odalric-Ambrym Maillard\thanks{\begin{minipage}[t]{\linewidth}Université de Lille, Inria, CNRS, Centrale Lille UMR 9189 – CRIStAL, F-59000 Lille, France\\
    \url{odalric.maillard@inria.fr}\end{minipage}}
    \and David Emukpere\thanks{\begin{minipage}[t]{\linewidth} Work done at Inria, F-59650 Villeneuve d'Ascq, France\\
    \url{}\end{minipage}}
}
\authorhead{Gautron \& \textit{al.\@}}
\RRtitle{gym-DSSAT : un modèle de cultures converti en un environnement d'apprentissage par renforcement}
\RRetitle{gym-DSSAT: a crop model turned into a Reinforcement Learning environment}
\titlehead{\texttt{gym-DSSAT} RL environment}
\RRresume{La résolution d'un problème de décision séquentielle en conditions réelles s'appuie très souvent sur l'utilisation d'un simulateur qui reproduit ces conditions réelles. Nous introduisons un nouvel environnement pour l'apprentissage par renforcement (AR) qui propose des tâches d'apprentissage réalistes pour la conduite de cultures. \texttt{gym-DSSAT} est une interface \texttt{gym} avec le simulateur de cultures Decision Support System for Agrotechnology Transfer (\texttt{DSSAT}), un simulateur de haute fidélité. \texttt{DSSAT} a été développé durant les 30 dernières années et est largement reconnu par les agronomes. \texttt{gym-DSSAT} propose des simulations prédéfinies, basées sur des expérimentations au champ avec du maïs. L'environnement est aussi simple à utiliser que n'importe quel autre environnement \texttt{gym}. Nous proposons des performances de base dans l'environnement en utilisant des algorithmes d'AR conventionnels. Nous décrivons également brièvement comment le simulateur monolithique DSSAT, codé en Fortran, a été transformé en un environnement d'AR en Python. Notre approche est générique et peut être appliquée à des simulateurs similaires. Quoique très préliminaires, les premiers résultats expérimentaux indiquent que l'AR peut aider les chercheurs à rendre les pratiques de fertilisation et d'irrigation plus durables.
}
\RRabstract{Addressing a real world sequential decision problem with Reinforcement Learning (RL) usually starts with the use of a simulated environment that mimics real conditions. We present a novel open source RL environment for realistic crop management tasks. \texttt{gym-DSSAT} is a \texttt{gym} interface to the Decision Support System for Agrotechnology Transfer (\texttt{DSSAT}), a high fidelity crop simulator. \texttt{DSSAT} has been developped over the last 30 years and is widely recognized by agronomists. \texttt{gym-DSSAT} comes with predefined simulations based on real world maize experiments. The environment is as easy to use as any \texttt{gym} environment. We provide performance baselines using basic RL algorithms. We also briefly outline how the monolithic DSSAT simulator written in Fortran has been turned into a Python RL environment. Our methodology is generic and may be applied to similar simulators. We report on very preliminary experimental results which suggest that RL can help researchers to improve sustainability of fertilization and irrigation practices.
}

\RRmotcle{itinéraire technique, conduite des cultures, modèle de culture, agriculture, Apprentissage par Renforcement, DSSAT, OpenAI gym, Python}
\RRkeyword{crop management, crop model, agriculture, Reinforcement Learning, DSSAT, OpenAI gym, Python}

\RRprojet{Scool}

\RCLille

\begin{document}
\RRNo{9460}
\makeRR
\tableofcontents

\section*{Software availability}
\noindent \texttt{gym-DSSAT} [\url{https://gitlab.inria.fr/rgautron/gym_dssat_pdi}] is an open source software, released under a 3-Clause BSD licence. A complete documentation is available [\url{https://rgautron.gitlabpages.inria.fr/gym-dssat-docs/}]. \texttt{gym-DSSAT} uses a modification of the Decision Support System for Agrotechnology Transfer (\texttt{DSSAT}) software (\url{https://dssat.net/}) and the PDI Data Interface (\texttt{PDI}) library (\url{https://pdi.dev/master/}). Both \texttt{DSSAT} and \texttt{PDI} are open source software, released under a 3-Clause BSD licence. In this work, we used \texttt{gym-DSSAT  0.0.7}.

\section{Introduction}
\label{sec:introductionGymDssat}
During a growing season, farmers perform series of crop operations in their fields in order to reach production objectives. They make these decisions under uncertainty, for instance weather uncertainty. We consistently use the adjective uncertain for events with unsure realizations.
%Thereby, crop management is inherently a sequential decision problem under uncertainty.
Reinforcement Learning (RL) addresses such problems where an agent learns to control the evolution of an unknown and uncertain dynamical system, in order to perform a given task. In RL, addressing a complex real-world problem usually starts with the use of a high-fidelity simulator which mimics real learning conditions. We present \texttt{gym-DSSAT}, an RL environment based on a celebrated crop model, the Decision Support System for Agrotechnology Transfer \citep[\texttt{DSSAT},][]{hoogenboom2019dssat} cropping system model. In this introduction, we define the concepts of crop management, mechanistic models and RL, and show how \texttt{gym-DSSAT} ties together these notions as an RL environment for crop management tasks.

Crop management is the series of crop operations a farmer performs in a field in order to reach production objectives \citep{sebillotte1974agronomie, sebillotte1978itineraires}, such as reaching at least minimum yield and grain protein content. In a field, complex physical, chemical and biological dynamical processes interact \citep{Husson2021}. Uncertain factors, such as weather events, drive the evolution of this dynamical system. In rainfed cropping systems, i.e.\@ non-irrigated cropping systems, rainfall is a major determinant of maize yield  besides nutrient availability \citep{mueller2012closing, kadam2014agronomic, li2019excessive}. Water stress occurring during maize flowering period may greatly reduce final grain yield \citep{kamara2003influence}. Weather forecasts remain highly uncertain beyond 1-month lead time \citep{hao2018seasonal}. Consequently, at the beginning of the growing season, harvest is highly uncertain in rainfed cropping systems.  %Consequently, at the time decisions are taken, crop operations result in inherently uncertain outcomes. %During a growing season, which generally lasts from three to six months, all crop operations jointly impact a field plot evolution. A farmer may observe effects of a crop operation after a long delay. As an example, an uneven maize sowing depth is likely to lead to an uneven emergence of plants which later induces a partial sterility of the plant population due to the competition of dominant individuals which have first emerged. Thereby, an improper sowing may strongly affect the final grain yield in an irreversible manner. 

Learning sustainable crop management practices is not a trivial task. Nitrogen fertilization requires future minimum rainfall and temperature following the application for the fertilized nitrogen to become available to plants. For an efficient nitrogen fertilizer management, available nitrogen in soil must match plant uptake, both in time and quantity \citep{meisinger2002principles}. %Nitrogen availability is not immediate after fertilization. 
Indigenous soil nitrogen supply, i.e. nitrogen supply which does not come from fertilizer applications during the current growing season, is often the first crop nitrogen supplier \citep{cassman2002agroecosystems}. If total nitrogen supply is greater than total plant uptake, the excess of nitrogen will be a source of water pollution, especially with excessive rainfall. If total nitrogen supply is less than total plant nitrogen uptake, then crops may suffer nitrogen deficiency. Maize nitrogen uptake depends on growth stage, and is greater during silking \citep{hanway1963growth}. Early and severe maize nitrogen deficiencies require earlier nitrogen supply compared to situations without such early nitrogen deficiencies \citep{binder2000maize}.
%As an example, droughts affect maize anthesis \citep[e.g.\@ ][]{herrero1981drought} where nitrogen uptake is maximum. 
Thereby, designing an optimal fertilization policy is a complex task. At the time a farmer makes a decision on fertilization, future plant nitrogen uptake, temperature, rainfall and other important factors that determine nitrogen plant nutrition are uncertain and so are the consequences of nitrogen applications \citep{morris2018strengths}.

In order to address complex crop management decisions, such as designing fertilization or irrigation policies, scientists have developed specialized simulators. Mechanistic models are based on the laws of nature and implemented with expert knowledge to simulate physical, chemical, and/or biological processes with high fidelity \citep[][]{sokolowski2012handbook}. These models have often evolved into complex software over decades of research and collaborative development. Crop models, often called process-based crop models, are mechanistic models which a user uses to simulate crop growth, generally at the plot scale. They model interactions among crops, soil, atmosphere, and crop operations \citep[e.g.\@ planting, fertilizing: see][]{wallach2018working}. As an example, the \href{https://dssat.net}{Decision Support System for Agrotechnology Transfer}\footnote{\url{https://dssat.net}} \citep[\texttt{DSSAT}, ][]{hoogenboom2019dssat} software is a high-fidelity crop model developed over the past three decades. \texttt{DSSAT} is widely recognized by agronomists for crop simulations. It is based on the daily integration of a set of partial differential equations describing the various processes at stake. For instance, nitrogen dynamics partially depend on soil dynamics (e.g.\@ mineralization processes or soil water flows) and plant uptake (itself partially determined by physiological processes such as carbohydrate allocation in plant, depending on growth stages). Crop models can be used as exploratory tools to find best management practices. For instance \citet{he2012identifying} identified best sweetcorn irrigation and fertilization practices in Florida, USA, based on simulations.
%For instance, \citep[][]{adam2020more} usinf the the Agricultural Production Systems sIMulator \citep[APSIM,][]{holzworth2014apsim} crop model showed that for the Sudano-Sahelian zone, the effect of a change in cultivar was marginal compared to an increase in nitrogen fertilizer rates for sorghum. % PP: I remove this sentence because I do not see its necessity here: For instance, simulators allow the pretraining an RL agent with Transfer Learning techniques \citep[e.g.][]{rusu2017sim} and cheap, realistic \textit{in silico} testing. 

\begin{figure}
    \centering
    \includegraphics[height=.2\textheight]{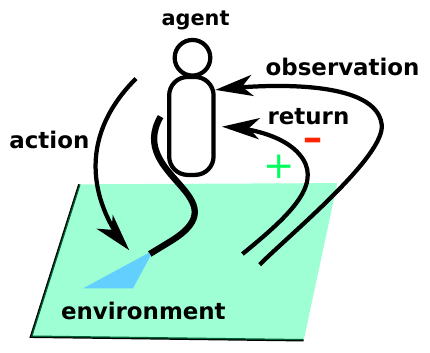}
    \caption{The Reinforcement Learning loop. The goal of the decision maker, called the agent, is to control the evolution of a dynamical system called the environment, in order to perform a given task. Sequentially, the agent observes the environment and takes an action based on this observation. The action affects the environment, and the agent receives a return that indicates how it is performing regarding the task to perform. The process repeats until the decision series eventually ends.}
    \label{fig:rlLoop}
\end{figure}

Reinforcement Learning \citep[RL,][]{sutton2018reinforcement} is a domain of Machine Learning (ML) and more generally Artificial Intelligence (AI) that addresses sequential decision problems under uncertainty. A decision maker, called an agent, interacts with a dynamical system called the environment which dynamics may be stochastic. The  goal of the agent is to control the evolution of the environment in order to perform a given task. Along a series of decisions, named an episode, the agent sequentially interacts with its environment until the decision sequence eventually ends. At each time step of an episode, the agent observes its environment, decides on an action and performs it. After the agent has taken an action, the action impacts the environment, and the agent receives a return from the environment. In general, the return is a scalar value, which indicates how the agent is performing regarding his task. The agent task is to maximize, in expectation, the total reward it has collected during an episode. To do so, the agent learns from multiple episodes in a trial and error fashion. Figure~\ref{fig:rlLoop} illustrates the interaction loop occurring during an episode. One can think of the ``hot and cold" kid game where the hunter's goal is to find a hidden object in a room. Each time after the hunter has moved, if he gets closer to its target, the other kids indicate ``hotter", else ``colder". Based on trial and error, the hunter will try to refine its position to maximize the temperature. This process repeats until the hunter finally finds the object, and the episode ends. This simple example illustrates the concepts of RL, where the hunter is the agent, the environment is the room with the position of the hunter and the hidden object, and finally the temperature is the return. RL generalizes these concepts to the stochastic case where after each action, the environment evolution and returns are drawn from probability distributions. RL seems an relevant tool to solve crop management problems, and in particular, to address sustainable agriculture challenges \citep{binasreinforcement, gautron2022reinforcement}.

In the vast majority of RL applications, researchers only experiment with simulated RL environments. Nonetheless, RL algorithms ultimately intend to directly learn from real-world interactions \citep[][Chapter 17, Section 6]{sutton2018reinforcement}. Still, real-world RL applications generally begin with the use of a simulator of the environment as testbed for candidate algorithms, and/or used to facilitate real-world learning with the help of prior knowledge learned from simulated interactions. In the latter case, such knowledge transfer from imperfect simulations to reality is still challenging in practice \citep[e.g.][]{golemo2018sim}. The simulation of real conditions require complex models that accurately mimic the evolution of the environment. These simulators embed state-of-the-art and continuously evolving knowledge. Crop models are consequently of great interest to address real world crop management problems with RL.

Crop modellers historically belong to scientific communities that are generally far from the ML/RL communities. Crop models were not designed to fit into an RL interaction loop. Most of widely used crop models \citep[see examples in][]{camargo2016six} internally work on a daily state update during the growing season but do not allow daily interactions with the user (be it human or virtual). A user first parametrizes a simulation, which often requires substantial domain specific knowledge. Then the simulator runs until reaching a final state which is generally crop maturity. After completion of the simulation, the user accesses partial in-season intermediate and final field states that have been internally stored during the execution. Moreover, crop modelers usually have implemented these models using Fortran, C, or C++ programming languages whereas RL researchers tend to favor Python nowadays. It follows that turning a crop simulator into a proper RL environment --without the burden of simulation setting requiring advanced expert knowledge-- is challenging. To turn the monolithic \texttt{DSSAT} Fortran crop model into an RL environment, we introduce the use of the \href{https://pdi.dev}{PDI Data Interface}\footnote{\url{https://pdi.dev}} (\texttt{PDI}) which allows loose coupling between C/C++/Fortran code and Python code.
%In this work, we use \texttt{PDI} to daily pause DSSAT during its in-season execution, to read its internal state to provide an observation to the agent, to specify to DSSAT the action(s) the agent has taken for the current day and finally to resume DSSAT execution. 
Beyond \texttt{DSSAT}, this approach may be used to turn other C/C++/Fortran monolithic mechanistic models into RL environments. We think this approach could reveal the value of many existing simulators as RL environments.

%This report is written considering readers who belong to the reinforcement learning community. However, we also hope that this report will also draw the interest of agronomists, or more generally people developing mechanisctic models, who are not familiar with reinforcement learning. With them in mind, we introduce the notions of RL that are necessary to know to read this paper. This report is organized as follows. 
Section~\ref{sec:relatedWork} presents similar works which turned crop models into RL environments. Section~\ref{sec:backgroundOnRL} briefly introduces mathematical and practical formalization of RL problems. Section~\ref{sec:decisionProblem} describes \texttt{gym-DSSAT} features and decision problems. In Section~\ref{sec:structure}, we show the internals of \texttt{gym-DSSAT} in a nutshell. Section~\ref{sec:baselines} provides an example of how to address the problem of maize nitrogen fertilization in \texttt{gym-DSSAT} as a use case, and discusses execution time and reproduciblity of experiments using \texttt{gym-DSSAT}. Finally, in Section~\ref{sec:futureWorks} we open on limits of our current crop management environment and discuss future improvements.

\section{Related work}
\label{sec:relatedWork}
Early seminal works addressed agricultural decision-making under uncertainty at the farm scale \citep{tintner1955stochastic, freund1956introduction}. The first case of an RL agent interacting with a crop simulator in order to learn crop management is found in \citet{garcia1999use}. The author used a modification of the \texttt{Déciblé} crop model \citep{chatelin2005decible}. The RL agent learned wheat sowing and nitrogen fertilization under pollution constraints. During simulations, weather series were stochastically generated. The modified version of \texttt{Déciblé} is not available anymore. In \citet{garcia1999use}, the RL agent did not manage to outperform the crop management policy of an expert. 
%Since then, very few works addressed crop management with RL \citep[e.g.\@]{trepos2014apprentissage}. 
Opportunities modern RL techniques bring for learning sustainable crop intensification have been prospected by \cite{binasreinforcement, gautron2022reinforcement}. Recently, several works directly used crop models or surrogate models as RL environments \citep[e.g.][]{sun2017reinforcement,wang2020deep, chen2021reinforcement}. However, none of these works has provided an open source and standardized crop management RL environment.

\cite{overweg2021cropgym} proposed \texttt{CropGym}, a \texttt{gym} interface to train an agent to perform wheat nitrogen fertilization. The environment uses the \href{https://pcse.readthedocs.io/en/stable/}{Python Crop Simulation Environment (PCSE)} \texttt{LINTUL3} \citep{shibu2010lintul3} wheat crop model.  Fertilization is treated as a weekly choice of a discrete amount of fertilizer to apply. The authors successfully trained an RL algorithm to address nitrogen fertilization in their RL environment. The agent performed better than the two expert fertilization policies they considered. In the aforementioned RL environment, there is no built-in stochastic weather generation. Overfitting describes the fact an algorithm, after being trained, performs poorly in unseen situations, despite having shown good performance in training situations. %Consequently, overfitting is characterized be overly-optimistic performance at the training stage. 
In \texttt{CropGym}, simulations use a limited set of historical weather records, which may favor overfitting due to limited randomness, especially for data intensive algorithms used in deep RL (see Section~\ref{sec:mdp}). %\cite{hitti2021growspace} proposed \texttt{GrowSpace}, an RL environment to learn to shape plants by controlling a moving a light source. Simulations are based on a plant light competition model.

\paragraph{Contribution}\texttt{gym-DSSAT} provides both maize fertilization and irrigation RL problems. Our RL environment features a built-in stochastic weather generator. We designed \texttt{gym-DSSAT} to allow researchers to easily customize realistic crop simulations of one of the most celebrated crop simulator, the \texttt{DSSAT} crop model. \texttt{DSSAT} datasets being abundant in the literature, \texttt{gym-DSSAT} allows to mimic a wide range of real-world growing conditions. Our Python RL crop management environment provides to the user a simple standardized interface, and still results in a lightweight software. Our technical approach is generalizable to any of the 41 other crops \texttt{DSSAT} simulates, and more broadly to other C/C++/Fortan mechanistic models.

\section{Formalization of RL decision problems}
\label{sec:backgroundOnRL}
\noindent Sections~\ref{sec:mdp} and~\ref{sec:pomdp} present most common mathematical formalization of RL decision problems. Section~\ref{sec:gym} presents \texttt{gym}, a practical pythonic interface to RL environments. 

\subsection{From Markov decision processes to reinforcement learning}
\label{sec:mdp}
\noindent Though RL paradigm may address a wide range of sequential decision problems, RL is usually employed to solve Markov Decision Problems (MDP). We introduce minimal materials on MDPs, for the reader to get an appropriate understanding of this paper. For an in-depth presentation of MDPs, see \citet{puterman1994markov}. 

\begin{figure}
    \centering
    \includegraphics[width=.5\textwidth]{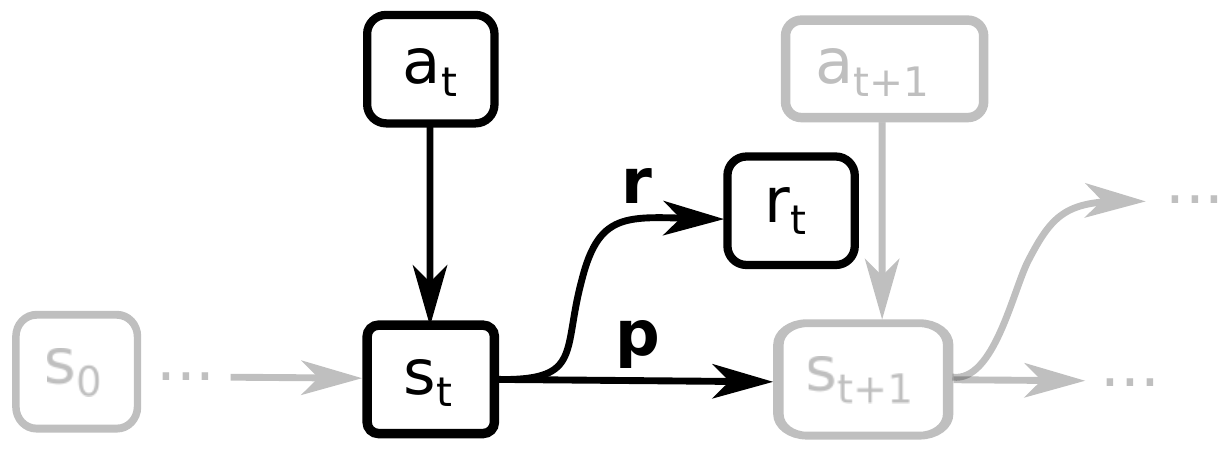}
    \caption{In reinforcement learning, a Markov decision process models the environment. At each time step $t$, the agent observes the  environment current state $s_t$. Depending on $s_t$, the agent takes an action $a_t$ according to its policy. As a consequence of taking action $a_t$, the environment transits to next state $s_{t+1}$, depending on the transition function $\textbf{p}$, and the agent observes the return $r_t$ which depends on the return function $\textbf{r}$. This process repeats until the episode eventually ends.}
    \label{fig:MDP}
\end{figure}

\paragraph{Markov decision process}%Before defining Markov decision problems, let us define Markov Decision (MD) processes. 
A Markov Decision (MD) process describes the evolution of a dynamical system over discrete time. The system evolution is impacted by the actions an agent can perform. % in which decisions are involved. %Though extensions to continuous time exist, we will consider such processes evolving in a discrete time.
An MD process $\mathcal{M}$ is defined by a tuple $\mathcal{M} = \langle \cS,\cA,{\bf p}, {\bf r}\rangle$.
At each decision step $t \in \{1, 2, 3, \cdots\}$, an agent observes the state of the environment $s_t\! \in\! \cS$ and takes an action $a_t\! \in \!\cA$, where $\cS$ is the state space, i.e. the set of all possible states and $\cA$ is the action space, i.e. the set of all possible actions. 
%Both state and action spaces may either be discrete or continuous.
Each action $a\in \cA$ leads to a stochastic transition from current state $s_t$ to next state $s_{t+1}$.
${\bf p}$, the transition function, defines the transition dynamics: ${\bf p} (s, a, s')$ is the probability the environment transits to state $s'$ if action $a$ has been performed in state $s$.
After performing an action, the agent receives a return, or reward, from the environment. Returns are given by the real function ${\bf r}$, named return function. ${\bf r} (s, a, s')$ is the expected return when action $a$ is performed in state $s$ leading to next state $s'$. 
%The interactive decision process between the agent and the environment is illustrated in \figurename~\ref{fig:MDP}. 
The interaction between an agent and an MD process generates a sequence $s_0, a_0, r_0, s_1, a_1, r_1, s_2, a_2, r_2, etc.$, called an episode, as \figurename~\ref{fig:MDP} illustrates. An MD process verifies the Markov property: the probability law of $s_{t+1}$ is fully specified by the knowledge of the current state $s_t$ and the action performed in this state $a_t$ at time $t$ (and ${\cal M}$).
There may exist a subset of states $\mathcal{S}_{\text{final}} \subset \mathcal{S}$, called the set of final states, such that when the agent reaches a state $s \in \mathcal{S}_{\text{final}}$, the episode ends.
%A MDP hence defines a discrete time dynamical process in which an agent interacts with an environment through actions.

\paragraph{Markov decision problem}%Now, we define a Markov decision problem as
A Markov decision problem (MDP) is a Markov decision process in which the agent has to optimize a given objective function. Let us consider an MDP in which the agents performs a given number $T$ of interactions and let us define the objective function $J$ as:
\begin{equation}
  J(T) = \sum_{t=0}^{T-1} r_t,
  \label{eq:J}
\end{equation}
% \oam{This looks very weird, why not $\sum_{t=0}^{T-1} r_t$ ?}
where $r_t$ is the return collected by the agent at time step $t$. The state reached at time $T$ is a final state. The agent goal is to maximize $J(T)$. A policy $\pi: {\cal S} \to {\cal \mathcal{P}(\mathcal{A})}$ maps each state to a distribution over the set of actions $\mathcal{P}(\mathcal{A})$. A policy specifies which action the agent performs in any state. The objective function $J(T)$ depends on the returns the agent has collected between $t=0$ and $t=T-1$. Collected returns depend on the agent policy, consequently, $J(T)$ is a function of the agent policy. The more a policy maximizes $J(T)$, the better the policy is. 
%Thereby, the agent goal is to find a policy $\pi$ which maximizes as much as possible $J(T)$.
%The question is then about finding a good policy, or an optimal one if such a policy does exist and may be computed.
Considering a policy $\pi$, we define the value of a state $s$ 
%when the agent follows a policy $\pi$ 
as the expectation of the objective function when the agent follows policy $\pi$ starting from state $s$:
\begin{equation}
    V_{\pi}(s) = \mathbb{E}_{\pi}\bigg[\sum_{t=0}^{T-1} r_{t}\bigg\vert s_0=s\bigg], \forall s \in \cS.
\label{eq:valueFunction}
\end{equation}
The Q-value of state $s$ and action $a$
%when the agent follows a policy $\pi$ 
is defined as the expectation of the objective function when the agent performs $a$ in $s$ and then follows $\pi$:
\begin{equation}
    Q_{\pi}(s, a) = \mathbb{E}_{\pi}\bigg[\sum_{t=0}^{T-1} r_{t}\bigg\vert s_0=s, a_0=a \bigg], \forall (s, a) \in \cS \times \cA.
%\label{eq:valueFunction}
\end{equation}

The goal of the agent is to learn an optimal policy $\pi^{\star}$ that maximizes the value in all states. For the MDPs we consider in this paper, it can be proven that at least one optimal policy exists \citep{puterman1994markov}.
%There may be more than one such optimal policy. 
When an MDP is fully defined, i.e.\@ $\cS$, $\cA$, ${\bf p}, {\bf r}, ~\text{and}~ T$ are known to the agent, finding an optimal policy is an optimization problem
%known as a planning problem % may be confusing for planning when you just have a simulator but no analytic expression of p and r
where all necessary quantities to compute a solution are available. For instance, dynamic programming can be employed to approximate an optimal policy. When ${\bf p}, {\bf r}$ (and $T$) are unknown, then RL can be employed. In the latter case, in general ${\bf p}$ and ${\bf r}$ can only be sampled through interactions of the agent with the environment. 
%Various algorithms have been proposed to solve this problem and learn an optimal policy. 
Most of RL algorithms belong to one of the three following families \citep{sutton2018reinforcement}: (1) \textit{critic} methods which are algorithms that learn a value function (e.g.\@ Q-Learning, FQI, DQN) and then derive an optimal policy from it; (2) \textit{actor} methods which are algorithms that directly learn an optimal policy (e.g.\@ REINFORCE); (3) \textit{actor-critic} methods which simultaneously combine actor and critic methods (e.g.\@ A2C, PPO, SAC). In order to deal with potentially very large state and/or action spaces, RL algorithms generally use function approximators, to compactly represent value and/or policy functions. Deep RL is a special case of RL where function approximators are neural networks \citep{lapan2018deep}.

\subsection{Partially Observable Markov Decision Process}
\label{sec:pomdp}
An MDP is an idealized model of a real-world system because real systems are unlikely to verify the properties associated to MDPs, in particular the Markov property. A field plot involves many interleaved dynamical processes, and parameters which are still partially discovered/measurable and the study of these dynamics are active areas of research \citep[e.g.][]{Husson2021}. In an MDP, each state is supposed to contain all necessary information for the agent to be able to decide which action is the best to perform in order to optimize the objective function. %, and thus to learn an optimal policy. 
Except from synthetic problems like games with complete information, such as the game of Go, for most systems, the exact environment state is unknown to the agent. In contrast, with real-world systems, the agent is likely to only access uncertain or incomplete observations of states. Such problems can be formalized as Partially Observable Markov Decision Problem \citep[POMDP, ][]{aastrom1965optimal}. POMDPs are a specific topic of study in the RL literature, and require \textit{ad-hoc} algorithms to solve them \citep[e.g.\@][]{spaan2012partially}. %Other formalisms exist, such as Predictive State Representation \citep[PSR,][Chapter 17, Section 3]{sutton2018reinforcement} which guarantees a more compact representation than POMDP.
%As we will see in Section~\ref{subsec:defaultTasks}, \texttt{gym-DSSAT}'s problems are POMDPs. %In a POMDP, one solution to recover a notion of state consists in considering the history of all interactions of the agent with it environment. 
%MDPs provide limited modelization. Extensions can be found for continuous time, action or state space \citep[e.g.][]{van2007reinforcement, guo2009continuous}. 

%%%%%%%%%%%%%%%%%%%%%%%%%%%%%%%
%%%%%%%%%%%%%%%%%%%%%%%%%%%%%%%
\subsection{\texttt{gym} environments}
\label{sec:gym}
\begin{figure}
  \centering
  \includegraphics[width=.6\textwidth]{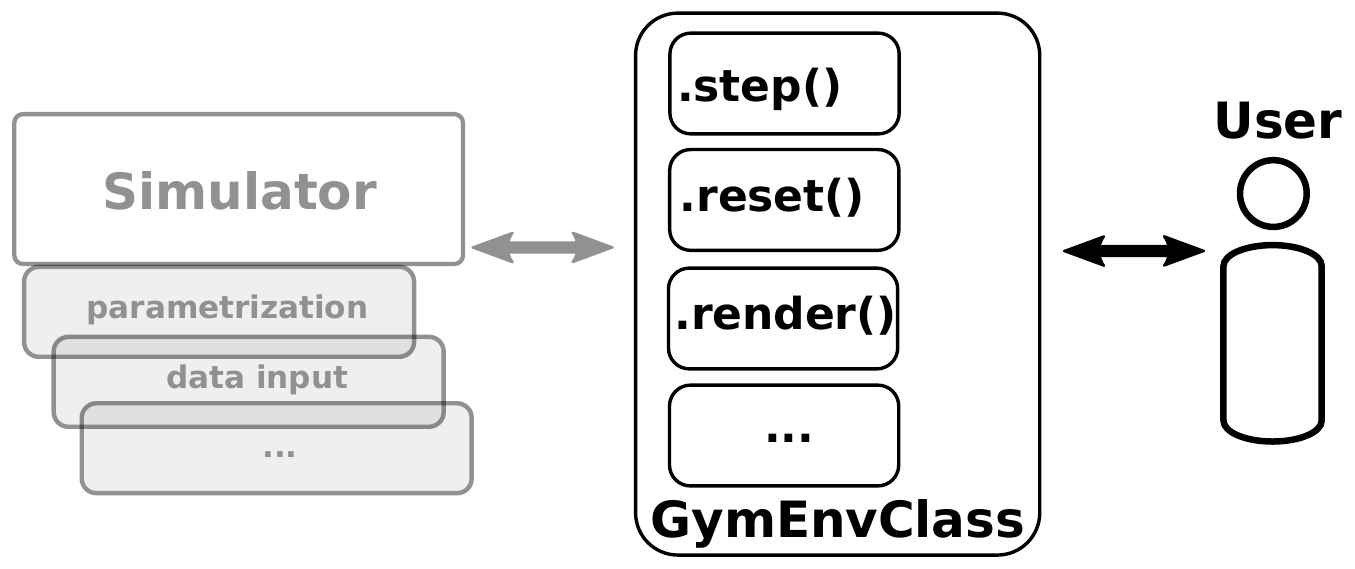}
  \caption{From a user's perspective, \texttt{gym} environments are simplified interfaces to simulators, through standardized methods.}
  \label{fig:gymStruct}
\end{figure}

\href{https://gym.openai.com/}{OpenAI gym}\footnote{\url{https://gym.openai.com/}} is an open source toolkit initially developed by the Open AI company, that provides light RL environments with a standardized Application Programming Interface (API). %OpenAI gym, gym for short,
\texttt{gym} API has become a reference in the RL community to create standardized RL environments in order to compare performances of RL algorithms. Many environments are available with \texttt{gym}, for instance with simulated games or physical dynamical systems, including robots. %Realistic \texttt{gym} environments are increasingly developed, such as with \citet{highway-env} for autonomous driving.
Typically, \texttt{gym} environments are straightforward to use: all simulated dynamics are pre-parametrized and hidden. The user instantiates an environment as simply as:
\begin{python}
import gym
env = gym.make("CartPole-v0") # create an instance of the environment CartPole-v0
\end{python}
As \figurename~\ref{fig:gymStruct} shows, \texttt{gym} is a wrapper that gives access to %encapsulates
a more complex simulator. \texttt{gym} environments come with default attributes which specify action and observation spaces. For instance in the case of the \texttt{CartPole-v0} environment, the user gets the specifications of a four-dimensional state space and a set of two possible actions:
\begin{python}
>>> env.observation_space # outputs observation lower bound, upper bound, shape, data type
Box(-3.4028234663852886e+38, 3.4028234663852886e+38, (4,), float32)
>>> env.action_space # if Discrete class, outputs the number of possible values
Discrete(2)
\end{python}
%Gym allows to connect any Python policy to the environment: from simple hard-coded deterministic functions, to complex neural networks. 

The user interacts with the environment through standardized methods. \texttt{gym} is independent of the implementation of the agent policy. The agent interacts with the environment by calling the \texttt{step()} method with an argument $a_t$ specifying the action to take, in order to receive the transition and reward generated by ${\bf p}$ and ${\bf r}$. The objective function is neither part of \texttt{gym} implementation, hence to optimize Equation~\ref{eq:valueFunction}, one specifies $\gamma$ as a parameter of the learning algorithm trying to build an optimal policy.

To illustrate the simplicity of interactions, we exhibit a basic RL loop:
\begin{python}
observation = env.reset() # reset the environment and get initial observation
# >>> observation
# array([-0.03325944, -0.02851367,  0.00086817, -0.00618905])

done = False # True when the episode is ended
while not done:
   action = policy(observation)  # get action depending on agent policy
   observation, reward, done, info = env.step(action) # perform the action
   # update the policy
env.render() # graphical representation of environment state
env.close()  # gracefully exits the environment
\end{python}
% PP removes these 2 lines from this python code because I do not see how it is useful to the reader and it is not explained anywhere.
%   # >>> env.step(env.action_space.sample())
%   # (array([-0.0304978 , -0.02853673, -0.00522757, -0.00568033]), 1.0, False, {})
% and added "update the policy"
\noindent \texttt{observation} corresponds to a possibly incomplete MDP state $s_t$, \texttt{reward} corresponds to $r_t$, \texttt{done} is True if the episode has ended, i.e. if the agent has reached a final state, and finally \texttt{info} provides optional extra information about the environment. We refer to the documentation available at \url{https://gym.openai.com/} for further details.

\section{Decisions problems in \texttt{gym-DSSAT}}
\label{sec:decisionProblem}
\noindent In Section~\ref{subsec:defaultTasks} we introduce the default 
crop management problems \texttt{gym-DSSAT} provides and Section~\ref{sec:customTasks} outlines how a user can create customized crop management problems.

\subsection{Default crop management problems of \texttt{gym-DSSAT}}
\label{subsec:defaultTasks}
By default, \texttt{gym-DSSAT} sequential decision problems simulate a maize experiment which has been carried out in 1982 in the experimental farm of the University of Florida, Gainesville, USA \citep[][\texttt{UFGA8201} experiment]{hunt1998data}. An episode is a simulation of a growing season. A simulation starts prior to planting and ends at crop harvest which is automatically defined as the crop maturity date. Crop maturity, a final state in \texttt{gym-DSSAT}, depends on crop growth, which depends itself on crop management and weather events, and the time to reach it is stochastic. Note that other final states exist in \texttt{gym-DSSAT}. For instance, improper crop management or too stressing weather
conditions may lead to early crop failure, which is also a final state. During the whole growing season (about 160 days on average), an RL agent daily decides on the crop management action(s) to perform: fertilize and/or irrigate. By default, for each episode, the weather is generated by the \texttt{WGEN} stochastic weather simulator \citep{richardson1985weather, soltani2003statistical}. \texttt{WGEN} has been parametrized based on historical weather records of the location of the original experiment. The duration between the starting date of the simulation and the planting date, which lasts about one month, induces stochastic soil conditions at the time of planting (e.g. soil nitrate, or soil water content), as a result of stochastic weather events.

The number of measurable attributes in a field is extremely large. In contrast, farmers have been described to make crop management decisions based on a limited practicable number of field observations \citep{papy1998savoir}. For this reason, in \texttt{gym-DSSAT}, the RL agent only accesses a restricted subset of \texttt{DSSAT} state variables which constitutes the observation space of the environment. Based on agronomic knowledge, we selected this subset with the constraint of the variables to be realistically measurable/proxiable in real conditions. These observation variables are mixed, and take either continuous or discrete values. We documented all observation and action variables in the \texttt{gym-DSSAT} \href{https://gitlab.inria.fr/rgautron/gym\_dssat\_pdi/-/blob/stable/gym-dssat-pdi/gym\_dssat\_pdi/envs/configs/env\_config.yml}{YAML configuration file}\footnote{\url{https://gitlab.inria.fr/rgautron/gym\_dssat\_pdi/-/blob/stable/gym-dssat-pdi/gym\_dssat\_pdi/envs/configs/env\_config.yml}}. This file includes description of variables type, range, and agronomic meaning.

In \texttt{DSSAT}, the \texttt{WGEN} stochastic weather simulator is implemented as a first-order Markov chain, but all other processes are deterministic. Therefore, \texttt{gym-DSSAT} decision problems are Markovian. Because the agent only accesses a subset of all \texttt{DSSAT} internal variables, a \texttt{gym-DSSAT} problem is a POMDP, similarly to the real problems faced by farmers. From an RL perspective, one can rigorously address a \texttt{gym-DSSAT} decision problem as a POMDP, or follow the common pragmatic approach which treats a POMDP as an MDP. In contrast with many toy RL environments, the environment is autonomous: it evolves by itself and not only because an action has been performed by the agent. Indeed, if on a given day a farmer does not fertilize/irrigate, its field plot still evolves. A do-nothing action is always available at each time step, which corresponds to the spontaneous field evolution.
%Because we included zero fertilization and irrigation in the action space, the autonomous field evolution simply corresponds to a null action, which recovers an usual (PO)MDP.

\texttt{DSSAT} simulates dynamics at the plot level; likewise, the agent performs actions on the whole field plot. %In this section, we detail each crop management problem. 
Growing conditions such as soil characteristics and other crop operations such as soil tillage, cultivar choice %--except the weather and the action(s) to be performed--
are fixed. We defined default return functions based on agronomic knowledge following the reward shaping principle \citep{randlov1998learning,ng1999policy}, such that rewards were as much frequent and as much informative as possible regarding the desired behaviour of the agent. Reward shaping aims both at facilitating an agent learning and to steer policies towards desirable trade-offs such as maximizing grain yield and minimizing induced pollution.
%For formal definitions of return functions, please refer to the \href{https://gitlab.inria.fr/rgautron/gym\_dssat\_pdi/-/blob/stable/gym-dssat-pdi/gym\_dssat\_pdi/envs/configs/rewards.py}{python file of return definitions}\footnote{\url{https://gitlab.inria.fr/rgautron/gym\_dssat\_pdi/-/blob/stable/gym-dssat-pdi/gym\_dssat\_pdi/envs/configs/rewards.py}}.
We define return functions in a \href{https://gitlab.inria.fr/rgautron/gym\_dssat\_pdi/-/blob/stable/gym-dssat-pdi/gym\_dssat\_pdi/envs/configs/rewards.py}{standalone Python file},
and users can find admissible values of actions in the environment \href{https://gitlab.inria.fr/rgautron/gym\_dssat\_pdi/-/blob/stable/gym-dssat-pdi/gym\_dssat\_pdi/envs/configs/env\_config.yml}{YAML configuration file}, or in \texttt{gym-DSSAT} action space attribute.

By default, \texttt{gym-dssat} provides three RL problems:
\begin{enumerate}[nosep]

\item[\boxed{1}] A \textbf{fertilization problem} in which the agent can apply every day a real valued quantity of nitrogen, as indicated in \tablename~\ref{tab:actions}. Crops are rainfed, and no irrigation is applied during the growing season, excepted a single one before planting. \texttt{DSSAT} automatically performs planting operation when soil temperature and humidity lie in favorable ranges. Denoting $\texttt{trnu}(t,t+1)$ the plant nitrogen uptake (kg/ha) from its environment between days $t$ and $t\!+\!1$; and $\texttt{anfer}(t)$ the nitrogen fertilizer application (kg/ha) on day $t$, we crafted the default fertilization return function as:
\begin{equation}
  r(t) = \underbrace{\texttt{trnu}(t, t\!+\!1)}_{\substack{\text{plant nitrogen} \\ \text{uptake}}} - \underbrace{0.5}_{\substack{\text{penalty} \\ \text{factor}}} \times ~\underbrace{\texttt{anfer}(t)}_{\substack{\text{fertilizer} \\ \text{quantity}}}
  \label{eq:rewardFertilization}
\end{equation}
The return is the daily population nitrogen uptake (to be maximized) which we penalize if the agent has fertilized the previous day. We defined the penalty factor based on expert knowledge such that the return corresponds to a desirable trade-off between agronomic, economical and environmental potentially conflicting objectives. \tablename~\ref{tab:fertilizationObservationVariables} details the observation space.

\item[\boxed{2}] An \textbf{irrigation problem} in which the agent can provide every day a real valued quantity of water to irrigate, as indicated in \tablename~\ref{tab:actions}. Independently of agent actions, this problem features at the same time a deterministic low input nitrogen fertilization (see \tablename~\ref{tab:fertilizationExpertPolicy}). Planting date is fixed, about one month after the beginning of simulation. The daily-based return is the daily change in above the ground population biomass (to be maximized), which we penalize if the agent has irrigated the previous day, similarly to the fertilization problem. We provide default reward function in Appendix \figurename~\ref{eq:rewardIrrigation} and observation space in Appendix \tablename~\ref{tab:irrigationObservationVariables}.

\item[\boxed{3}] A mixed \textbf{fertilization and irrigation problem} which combines both previous decision problems: every day, the agent can fertilize and/or irrigate. Planting date is fixed, about one month after the beginning of simulation. In this case, the return has two components, one for each sub-problem: this is a multi-objective problem \citep[e.g.][]{hayes2021practical}. The default observation space is the union of the observation spaces of the fertilization and irrigation problems.
\end{enumerate}

\begin{table}
  \begin{center}
    \begin{tabular}{lll}
      \textbf{action}        & \textbf{description}                                 & \textbf{range}       \\ \hline
      fertilization & daily nitrogen fertilization amount (\SI{}{\kg}/\SI{}{\hectare}) & [0,200] \\
      irrigation    & daily irrigation amount (\SI{}{\liter}/\SI{}{\metre\squared})              & [0,50]
    \end{tabular}
    \caption{\texttt{gym-DSSAT} available actions}
    \label{tab:actions}
  \end{center}
\end{table}

\begin{table}\centering
  \begin{tabular}{cl}
    \toprule
    & \textbf{definition} \\
    \texttt{istage} & \texttt{DSSAT} maize growing stage \\
    \texttt{vstage} & vegetative growth stage (number of leaves) \\
    \texttt{topwt} & above the ground population biomass (kg/ha) \\
    \texttt{grnwt} & grain weight dry matter (kg/ha) \\
    \texttt{swfac} & index of plant water stress (unitless) \\
    \texttt{nstres} & index of plant nitrogen stress (unitless) \\
    \texttt{xlai} & plant population leaf area index (\SI{}{\metre\squared} leaf/\SI{}{\metre\squared} soil) \\
    \texttt{dtt} & growing degree days for current day (\SI{}{\celsius}/day) \\    
    \texttt{dap} & days after planting (day) \\
    \texttt{cumsumfert} & cumulative nitrogen fertilizer applications (kg/ha) \\
    \texttt{rain} & rainfall for the current day (L/\SI{}{\metre\squared}/day) \\
    \texttt{ep} & actual plant transpiration rate (L/\SI{}{\metre\squared}/day) \\
\bottomrule
\end{tabular}
\caption{Default observation space for the fertilization task.}
\label{tab:fertilizationObservationVariables}
\end{table}

\begin{table}\centering
  \begin{tabular}{cc}
    \toprule
    \textbf{DAP} & \textbf{quantity (kg N/ha)} \\
    40 & 27 \\
    45 & 35 \\
    80 & 54 \\
\bottomrule
\end{tabular}
\caption{Expert fertilization policy. `DAP' stands for Day After Planting.}
\label{tab:fertilizationExpertPolicy}
\end{table}

We did not define returns of decision problems as economic returns to avoid issues due to cost variations over time (e.g. petrochemicals). Fossil fuel necessary to produce artificial nitrogen fertilizers \citep[see the Haber process][]{modak2002haber} or to pump water are highly variable over time, making these decision problems non-stationary. Consequently, optimal solutions are likely to change through time. This is why we chose an arbitrary penalization of actions as a proxy of a notion of cost with sound agronomic trade-off, as shown in Equation~\ref{eq:rewardFertilization}. Despite their apparent simplicity, from an agronomic perspective, the three aforementioned decision problems are non-trivial. These problems can be made harder by providing a more restricted and/or noisy observation space to the agent, see the discussion of the fertilization use case (Section~\ref{subsec:fertilizationUseCase}).

\subsection{Custom scenario definition}
\label{sec:customTasks}
\begin{figure}
  \centering
  \includegraphics[width=.4\linewidth]{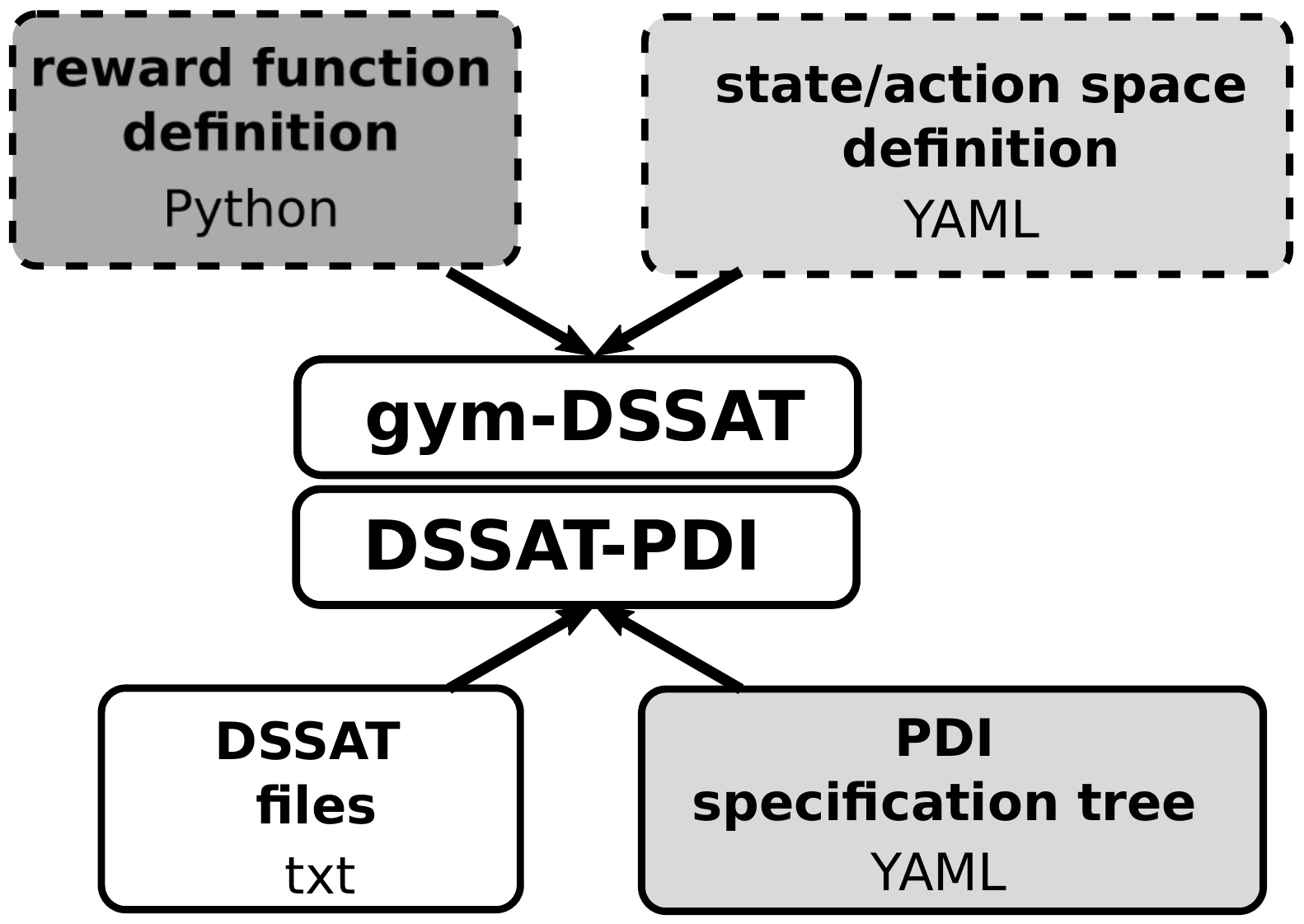}
  \caption{Configuration files used by the crop management reinforcement learning environment. At the top of the figure, files in dashed boxes define the reward function and state and action spaces of the Markov decision process. Dashed boxes indicate straightforward to customize configuration files. At the bottom of the figure, \texttt{DSSAT} files parametrize simulations, and the \texttt{PDI} specification tree is a technical file which manages the communication between \texttt{DSSAT-PDI} and \texttt{gym-DSSAT}.}
  \label{fig:config}
\end{figure}

\begin{figure}
  \centering
  \includegraphics[width=.6\linewidth]{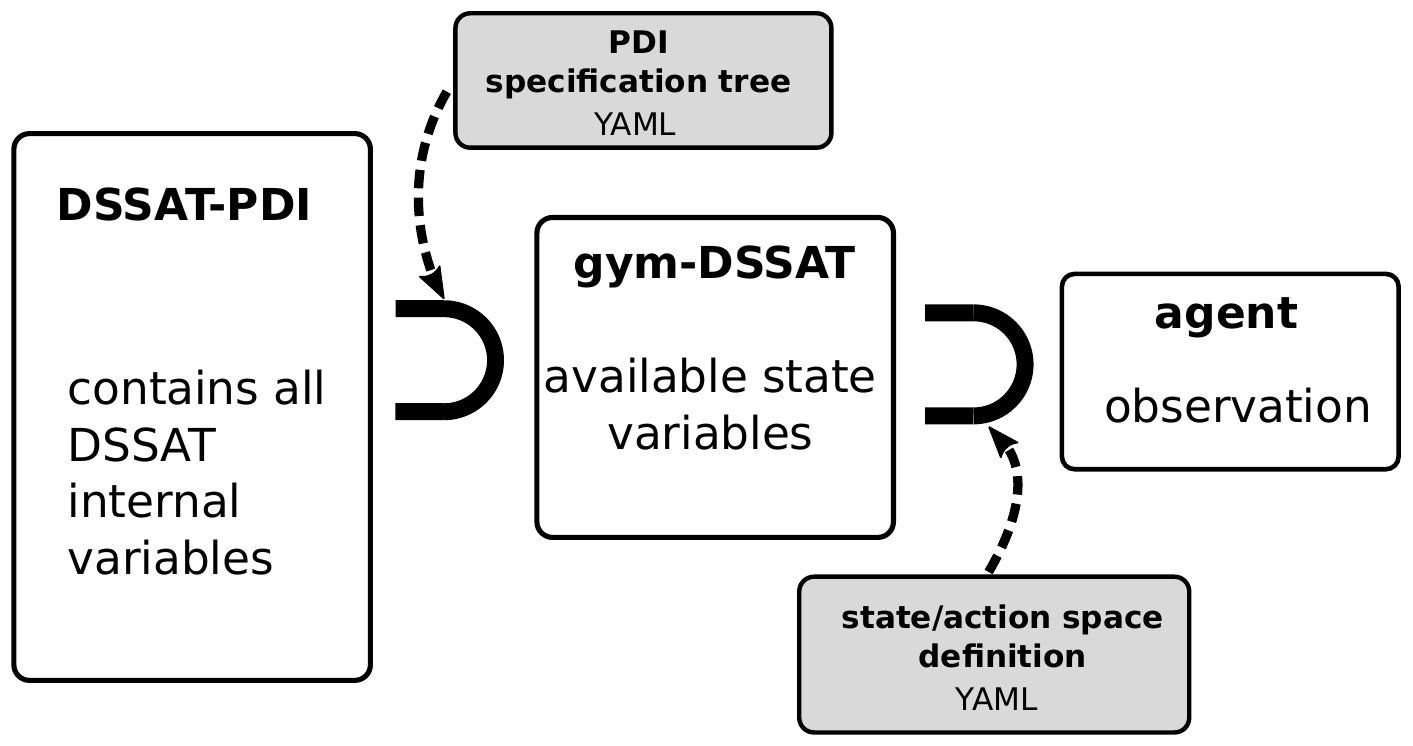}
  \caption{Successive subsets of \texttt{DSSAT} state variables until agent observations. Boxes filled with grey indicate files defining state variable subsets.}
  \label{fig:stateSpace}
\end{figure}

A user can customize \texttt{gym-DSSAT} problems, with an ease that depends on the features to be modified, see \figurename~\ref{fig:config}. An observation is a subset of \texttt{DSSAT} internal state variables. Figure~\ref{fig:stateSpace} shows the technical files which define the subset of variables constituting an observation. A user can straightforwardly modify the observation space in the \href{https://gitlab.inria.fr/rgautron/gym\_dssat\_pdi/-/blob/stable/gym-dssat-pdi/gym\_dssat\_pdi/envs/configs/env\_config.yml}{YAML configuration file}. In the same way, the definition of the return functions can be easily modified by the user by editing a \href{https://gitlab.inria.fr/rgautron/gym\_dssat\_pdi/-/blob/stable/gym-dssat-pdi/gym\_dssat\_pdi/envs/configs/rewards.py}{standalone Python file}\footnote{\url{https://gitlab.inria.fr/rgautron/gym\_dssat\_pdi/-/blob/stable/gym-dssat-pdi/gym\_dssat\_pdi/envs/configs/rewards.py}}. Built-in \texttt{DSSAT} features can be directly leveraged, such as environmental modifications with changes in atmospheric $\text{CO}_2$ concentration or meteorological features, to mimic the effects of climate change. Including other state variables, actions, crops, soil or weather generation parametrizations 
requires a deeper understanding of how \texttt{gym-DSSAT} works and some agronomic knowledge. This goes beyond the scope of this report; additional information is available in \texttt{gym-DSSAT} \href{https://gitlab.inria.fr/rgautron/gym\_dssat\_pdi/}{GitLab page}.

\section{Software architecture of the environment}
\label{sec:structure}
\paragraph{}In contrast with the simplicity of use of \texttt{gym-DSSAT}, we had to modify the original \texttt{DSSAT} simulator in a non-trivial manner to enable daily interactions with an agent and to interface the modified \texttt{DSSAT} Fortran program with Python. \texttt{DSSAT} was not designed to be used in an interaction loop. In this section, we detail how we have technically proceeded.

\subsection{The PDI Data Interface}

The PDI Data Interface \citep[\texttt{PDI},][]{roussel2017pdi} was the key element in \texttt{gym-DSSAT} which turned the original monolithic \texttt{DSSAT} simulator implemented in Fortran into an interactive Python RL environment. \texttt{PDI} is a library designed to decouple C/C++/Fortran codes, typically high-performance numerical simulations, from Input/Output (I/O) concerns. It offers a declarative low-invasive API to instrument the simulation source code, enabling the exposition of selected memory buffers used in the simulation to be read/written from/to \texttt{PDI}, and the notification to \texttt{PDI} of significant steps of the simulation. By itself, \texttt{PDI} does not provide any tool for the manipulation of data, instead it offers an event-driven plugin system to ease interfacing external tools with the simulation.

\texttt{PDI} moves most of the logic for the I/O interface away from the code: specifically, a YAML file is used to describe data structures and to specify when and which actions (provided by the different \texttt{PDI} plugins) to trigger on the selected data. The exposed data is selected by adding a few \texttt{PDI} calls in the source code with a very simple syntax. Other I/O libraries in the High Performance Computing field follow a similar declarative approach, such as \texttt{ADIOS-II}~\citep{godoy2020adios}, \texttt{Damaris}~\citep{dorier2016damaris} or \texttt{XIOS}~\citep{meurdesoif2013xios}. However, most of these alternatives are mainly focused on providing high-level abstractions of high-performace I/O operations, working with some domain-specific assumptions and providing additional features on top of parallel I/O streams, such as burst buffering or compression. \texttt{PDI} design has a general and global approach, aiming at more versatile scenarios, with a plugin system that enables substantially different possibilities and I/O strategies, such as the interaction with external Python code. As a result, \texttt{PDI} makes possible the implementation of \texttt{gym-DSSAT}: an external software (\texttt{gym}), directly interacts with a modification of a stand-alone, monolithic simulator (\texttt{DSSAT}).

% \JB{Maybe add a word on couplers, such as OASIS, OpenPALM or XIOS for weather \& climate models, there has to be others}

% \EP{I've re-written most of this subsection, removing some (IMHO) unnecessary low-level details about the operation with the different modern I/O frameworks available: ADIOS, Damaris, XIOS and PDI. I think those details are not specially relevant for this work, as the choice of PDI is mainly based on its versatility, that allow us the Python interaction.}

% \EP{Perhaps, we could extend the description about how PDI works, including a figure or some real code in listings (for example like Bruno and JB do in this paper using PDI with Dash: \url{https://hal.archives-ouvertes.fr/hal-03509198/file/HiPC__DEISA__Dask_Enabled_In_Situ_Analytics.pdf}), or maybe is enough with these two paragraphs, but I do not think we need more reasons/parameters to justify why we use PDI}

\begin{figure}
  \centering \subfloat[\centering \texttt{PDI} YAML file.]
  {{\includegraphics[width=.43\textwidth]{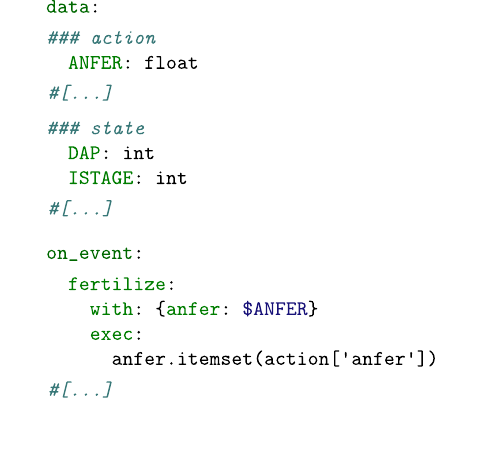}}
    \label{fig:yaml}}%
  \qquad \subfloat[\centering \texttt{DSSAT} code instrumented with \texttt{PDI}
  calls.]{{\includegraphics[width=.43\textwidth]{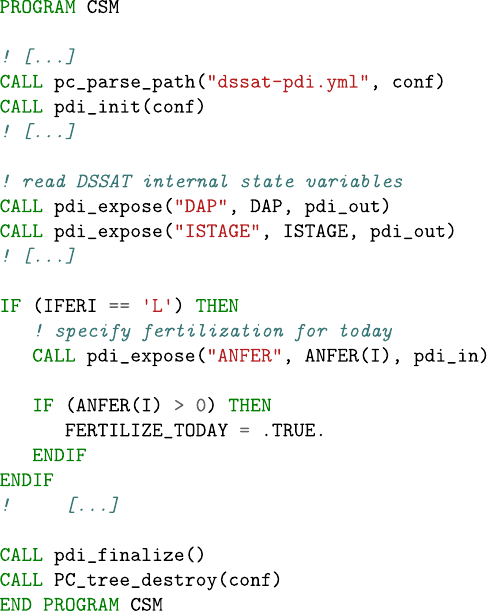}}\label{fig:instrcode}}%
  \caption{Simplified example of \texttt{PDI} use in \texttt{gym-DSSAT} for the fertilization decision problem. The left-hand side corresponds to the \texttt{PDI} specification tree (\figurename~\ref{fig:config}), and the right-hand side to the Fortran code of \texttt{DSSAT-PDI} (Section~\ref{sec:internals}). \label{fig:pdistuff}}%
\end{figure}

\figurename~\ref{fig:pdistuff} shows a simplified example of \texttt{PDI} use in
\texttt{gym-DSSAT}, for the fertilization problem. \figurename~\ref{fig:yaml} lists chunks of the YAML file with
declarations of exposed variables in the simulation code and
definitions of events to be triggered. This YAML file corresponds to the \texttt{PDI} specification tree file in \figurename~\ref{fig:config}. \figurename~\ref{fig:instrcode} shows a
snippet of the instrumented Fortran source code of \texttt{DSSAT}, with \texttt{PDI} initialization and three exposed simulation variables: two are read by \texttt{PDI} and will be available as observation variables, the third one is written by \texttt{PDI}, and corresponds to the action decided by the agent regarding crop fertilization for the current day. The whole instrumented code corresponds to \texttt{DSSAT-PDI}, see Section~\ref{sec:internals}.

\subsection{Internals of \texttt{gym-DSSAT}}
\label{sec:internals}
\begin{figure}
    \begin{center}
        \begin{subfigure}[t]{0.7\textwidth}
            \centering
            \includegraphics[width=\textwidth]{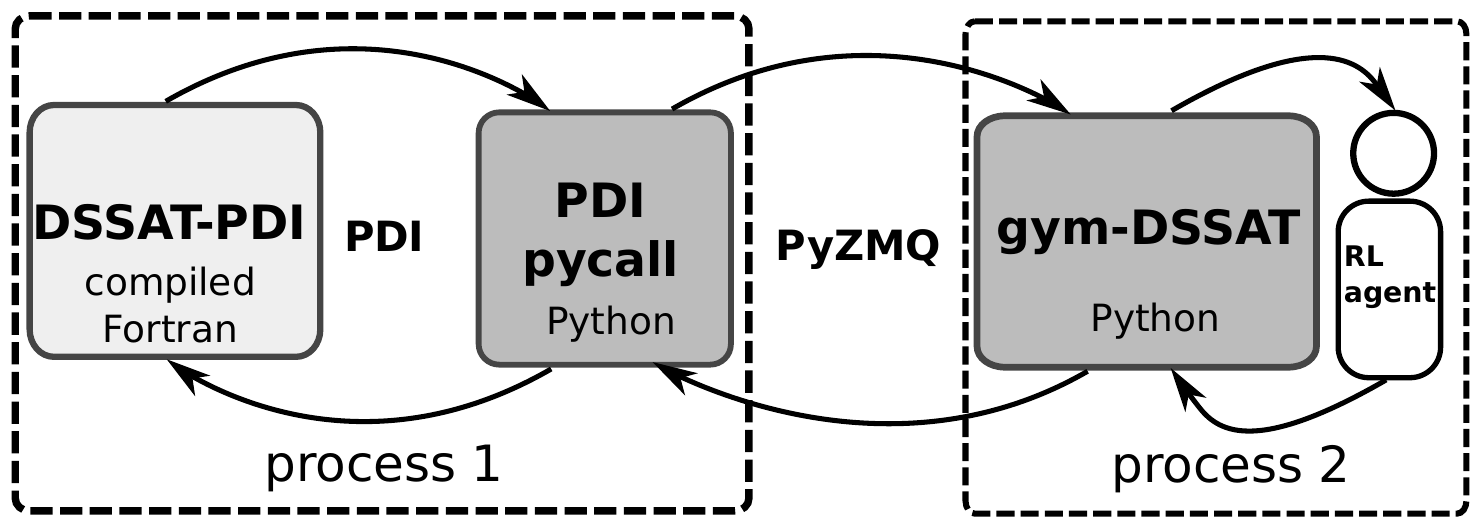}
            \caption{The reinforcement learning environment consists of two interacting processes. (i) the core modification of the \texttt{DSSAT} simulator, \texttt{DSSAT-PDI}, with its \texttt{PDI} module to execute Python code (\texttt{pycall plugin}); (ii) the \texttt{gym} Python interface \texttt{gym-DSSAT}. \texttt{PyZMQ} handles messaging between (i) and (ii) through Python sockets.}
            \label{fig:processes}
        \end{subfigure}
        \begin{subfigure}[t]{0.7\textwidth}
          \centering
          \includegraphics[width=\textwidth]{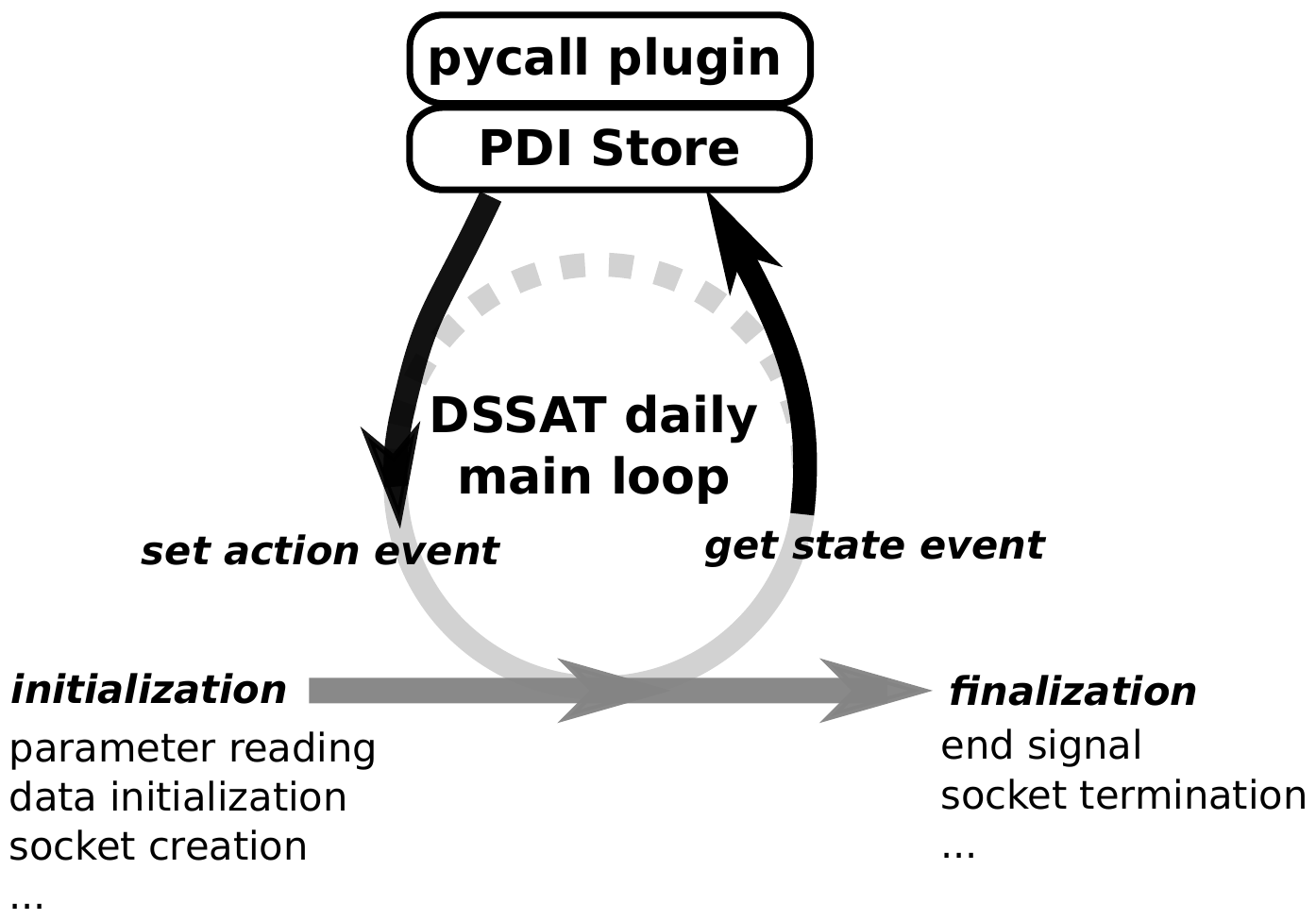}
          \caption{Simplified \texttt{PDI} data coupling and program execution of \texttt{DSSAT-PDI} which is the instrumented code of \texttt{DSSAT}. \texttt{PDI} handles the oftware initialization, data exchange with \texttt{gym-DSSAT} during the whole simulated growing season through the \texttt{pycall plugin}, and finally software graceful termination. The execution of the Python code by \texttt{PDI} \texttt{pycall plugin} is a blocking transaction.}
          \label{fig:dssatPdi}
        \end{subfigure}%
    \end{center}
    \caption{The elements of \texttt{gym-DSSAT}.}
    \label{fig:structure}
\end{figure}

\noindent We present a generic procedure which is an important methodological contribution of this work. \texttt{gym-DSSAT} is made of two communicating processes, as shows \figurename~\ref{fig:processes}:

\begin{itemize}[noitemsep,topsep=1pt]
\item[(i)] \texttt{DSSAT-PDI} which is the compiled Fortran code of a modification of the original \texttt{DSSAT} crop model, using the \texttt{PDI} library.
\item[(ii)] \texttt{gym-DSSAT} which, from a user perspective, is the usual \texttt{gym} interface to the RL environment.
\end{itemize}

\paragraph{DSSAT-PDI} The modification of the original \texttt{DSSAT} software, named \texttt{DSSAT-PDI}, allows an agent to daily interact with the crop simulator during a growing season. This interaction loop consists in repeatedly pausing \texttt{DSSAT}, reading \texttt{DSSAT} internal variables, providing these internal variables to the agent, specifying the action(s) of the agent to \texttt{DSSAT} and finally resuming \texttt{DSSAT} execution. \texttt{DSSAT} being in continuous development, the goal was to modify as little as possible the original source code for easy updates. Minimal interventions on \texttt{DSSAT} code have been facilitated by the \texttt{PDI} library. \texttt{PDI} %requires a few code snippets to be inserted in the source code to compile, and 
manages data communication with a Python process, through the \texttt{PDI pycall} plugin. \figurename~\ref{fig:dssatPdi} illustrates how \texttt{DSSAT-PDI} works. During the execution of the internal daily loop of \texttt{DSSAT}, \texttt{PDI} code snippets allow data coupling: accessing, writing in memory variables and triggering events. \texttt{DSSAT-PDI} execution starts with an initialization event, which provides all necessary elements for \texttt{PDI}, \texttt{DSSAT-PDI} and \texttt{gym-DSSAT} to start. Then, \texttt{DSSAT-PDI} enters its daily loop which consists in all successive daily updates of the crop simulator state during a growing season. While the daily loop executes, when the \texttt{get state} event occurs, \texttt{PDI} stores the values of a subset of \texttt{DSSAT} internal state variables in the \texttt{PDI} Store. After then, the \texttt{PDI} \texttt{pycall plugin} accesses these values, executes a Python script corresponding to the interaction with the agent, and stores back in the \texttt{PDI Store} the action(s) taken by the agent. Then, when the \texttt{set action} event occurs, \texttt{PDI} writes the variables corresponding to the agent action(s) into \texttt{DSSAT} memory and releases \texttt{DSSAT} daily loop execution. Finally, at the end of the simulation, a \texttt{finalization event} occurs to gracefully terminate the whole process. For the same parametrization and input, \texttt{DSSAT-PDI} and the vanilla \texttt{DSSAT} both consistently provide the same output.

\paragraph{gym-DSSAT} From a user's perspective, the \texttt{gym-DSSAT} environment is a simple interface to \texttt{DSSAT-PDI}, but from a technical point of view, \texttt{gym-DSSAT} handles all the execution of \texttt{DSSAT-PDI}. \texttt{gym-DSSAT} provides the necessary data input to \texttt{DSSAT-PDI}, including parametrization and weather data; it manages data communication and translation between the RL agent and \texttt{DSSAT-PDI} without extra effort. Finally,  \texttt{gym-DSSAT} is responsible for the graceful termination of \texttt{DSSAT-PDI}.

\paragraph{Messaging between \texttt{DSSAT-PDI} and \texttt{gym-DSSAT}.}
During the execution of a block of Python code, the \texttt{PDI} \texttt{pycall plugin} accesses \texttt{DSSAT-PDI} state variables, which have been previously stored in the \texttt{PDI Store}. Nevertheless, the data available in this Python process still requires to be communicated to \texttt{gym-DSSAT}, another Python process which is independent of the \texttt{DSSAT-PDI} process. As shown in \figurename~\ref{fig:processes}, the communication between \texttt{gym-DSSAT} and \texttt{DSSAT-PDI} is powered by ZeroMQ \citep{hintjens2013zeromq} Python sockets, with the \texttt{PyZMQ} package. Python sockets exchange data as JSON files, encoded as strings. Every transaction is a blocking event such that \texttt{DSSAT-PDI} daily loop is resumed only after \texttt{DSSAT-PDI} has received agent action(s).

\section{Experimenting with \texttt{gym-DSSAT}}
\label{sec:baselines}
\noindent In this section, we provide an RL use case for the maize fertilization problem using \texttt{gym-DSSAT}.
%RG: I think this is too informal % Our goal is really to demonstrate how a basic usage of gym-DSSAT is done, which results we get; it is in no way an investigation of the best performance that can be achieved by a reinforcement learning agent on gym-DSSAT: this is left as future work. 
We also discuss execution time and reproducibility issues using \texttt{gym-DSSAT}.

\subsection{Use case: learning an efficient maize fertilization}
\label{subsec:fertilizationUseCase}
\noindent As a simple use case, we present an example of how to address the fertilization task. We provide the irrigation use case in Appendix~\ref{sec:irrigationUseCase}. The source code of these experiments is available in \texttt{gym-DSSAT} \href{https://gitlab.inria.fr/rgautron/gym\_dssat\_pdi}{GitLab page}.

\paragraph{Methods}We consider the nitrogen fertilization task, as introduced in Section~\ref{subsec:defaultTasks}. The decision problem being on a finite horizon, i.e.\@ each episode lasted during a growing season, we defined the objective function of the agent as the undiscounted sum of returns, see Equation~\ref{eq:J}. \tablename~\ref{tab:fertilizationObservationVariables} presents the subset of \texttt{DSSAT} internal variables we have selected to define the observation space provided to the agent. These observation variables were selected as they could be realistically measured on farm. As a common practice, we pragmatically addressed this decision problem as an MDP, even though it is a POMDP (Section~\ref{sec:decisionProblem}). We used the Proximal Policy Optimization algorithm \citep[PPO,][]{schulman2017proximal}, as implemented in \texttt{Stable-Baslines3 1.4.0} \citep{stable-baselines}. PPO belongs to the family of deep RL actor-critic methods (see Section~\ref{sec:mdp}) and uses gradient descent to search for a good policy. PPO generally performs well on a wide range of problems and has been adopted as a standard baseline by the RL community. It is versatile as it can deal with both continuous and/or discrete actions and observation variables. %Actor-critic algorithms feature two separated learning entities, usually both implemented with a neural network. The first function corresponds to the agent policy which indicates how the agent acts depending on the environment state. This function is named \textit{actor}. The second function estimates the consequences of actions by approximating the values of states (see Equation~\ref{eq:valueFunction}). This function is named \textit{critic}. The critic evaluates how well the actor performs, in order to improve the performance of the policy.
\noindent In this experiment, we considered three policies:
\begin{itemize}
\item We first considered the most trivial fertilization policy: the ``null'' policy that never fertilized. As there still remains nitrogen in soil before cultivation \citep{morris2018strengths}, without mineral fertilization, the reference experiment, or \textit{control}, is usually the null policy. Agronomists then measure the effect of a nitrogen fertilization policy as a performance gain compared to the null policy, in order to decouple the effect of nitrogen fertilizer from the effect of already available nitrogen in soil \citep{vanlauwe2011agronomic}.
\item The second baseline is the `expert' policy, which is the fertilization policy of the original maize field experiment \citep[][\texttt{UFGA8201} experiment number \#1]{hunt1998data}, see Section~\ref{subsec:defaultTasks}. As \tablename~\ref{tab:fertilizationExpertPolicy} shows, this policy consists in three deterministic nitrogen fertilizer applications, which only depend on the number of days after planting.
\item Finally, the policy learned by PPO.
As our goal was not to obtain the best performance with an RL algorithm, but to simply establish a baseline, we used PPO default hyper-parameters as set in \texttt{Stable-Baselines3 1.4.0}. It is most likely better PPO hyper-parameters may be found. We trained PPO during ${10^6}$ episodes, with stochastic weather generation. The training procedure was light in terms of computation: it was possible to complete the ${10^6}$ episodes in about 1.5 hour of computation with a standard 8 core laptop. During training, the performance of PPO was evaluated on a validation environment every $10^3$ episodes. We seeded the validation environment with a different seed than for the training environment. Consequently, the validation environment generated a different sequence of weather series compared to the training environment. The model with the best validation performance was saved as the result of the training.
\end{itemize}

In order to compare fertilization policies, we measured their performance with $10^3$ episodes in a test environment. Test environment also featured stochastic weather generation, but with isolated seeds i.e.\@ different from the ones used in training and evaluation environments of PPO. This guaranteed that while testing policies, none of the stochastic weather series have been met by PPO during training or evaluation phases, in order to avoid over-optimistic performance measures \citep{stone1974cross}. In the performance analysis of policies, the evolution of returns $r_t$ provides information about the learning process from an RL perspective, but returns are still not directly interpretable from an agronomic perspective. Performance analysis of crop management options require multiple evaluation criteria \citep{dore2006agronomie, duru2015implement}. To remedy this problem, we used a subset of \texttt{DSSAT} internal state variables, provided in \tablename~\ref{tab:fertilizationTestingFeatures}, as performance indicators. Note that these variables are not necessarily contained in the observation space of the fertilization problem (\tablename~\ref{tab:fertilizationObservationVariables}) because we used them for another purpose than algorithm training. Each of these performance criteria covariates with the other ones. For instance, increasing the total fertilizer amount is likely to  increase the grain yield, but also likely to increase the pollution induced by nitrate leaching. The agronomic nitrogen use efficiency \citep[ANE, Equation 1,][]{vanlauwe2011agronomic} is a common indicator of fertilization sustainability. For a fertilization policy $\pi$, denoting $\texttt{grnwt}^{\pi}$ the dry matter grain yield of the policy $\pi$ (kg/ha), $\texttt{grnwt}^{0}$ the dry matter grain yield with no fertilization (kg/ha), and $\texttt{cumsumfert}^{\pi}$ the total fertilizer quantity applied with policy $\pi$ (kg/ha), we have:
\begin{equation}
        \text{ANE}^{\pi}(t)=\frac{\texttt{grnwt}^{\pi}(t) - \texttt{grnwt}^0(t)}{\texttt{cumsumfert}^{\pi}(t)}
        \label{eq:fertilizationEfficiency}
\end{equation}
The ANE indicates the grain yield response with respect to the null policy provided by each unit of nitrogen fertilizer.
Maximizing the ANE relates to economic and environmental aspects, leading to an efficient use of fertilizer which limits risks of pollution. Performance indicators presented in \tablename~\ref{tab:fertilizationTestingFeatures} express a complex trade-off between conflicting objectives.
%Maximizing the ANE is related to economic aspects \citep[benefit:cost ratio,][]{sassone1978cost} and environmental aspects (efficient use of fertilizer which limits risks of pollution). Consequently, the performance indicators presented in \tablename~\ref{tab:fertilizationTestingFeatures} expresses complex trade-offs between conflicting performance criteria.

\begin{table}\centering
  \begin{tabular}{cll}
    \toprule
 \textbf{variable}  & \textbf{definition} & \textbf{comment}       \\ %\cmidrule(lr){2-4}
\texttt{grnwt} & grain yield (kg/ha) & quantitative objective to be maximized \\
\texttt{pcngrn} & massic fraction of nitrogen in grains & qualitative objective to be maximized \\ 
\texttt{cumsumfert} & total fertilization (kg/ha) & cost to be minimized \\
-- & application number & cost to be minimized \\
-- & nitrogen use efficiency (kg/kg) & agronomic criteria to be maximized \\
\texttt{cleach} & nitrate leaching (kg/ha) & loss/pollution to be minimized \\
\bottomrule
\end{tabular}
\caption{Performance indicators for fertilization policies. An hyphen means \texttt{gym-DSSAT} does not directly provide the variable, but it can be easily derived.}
\label{tab:fertilizationTestingFeatures}
\end{table}

\begin{figure}%
    \centering
    \subfloat[\centering Mean cumulated return of each of the 3 policies against the day of the simulation. Shaded area displays the $\lbrack{}0.05, 0.95\rbrack$ quantile range for each policy.]{{\includegraphics[width=.47\textwidth]{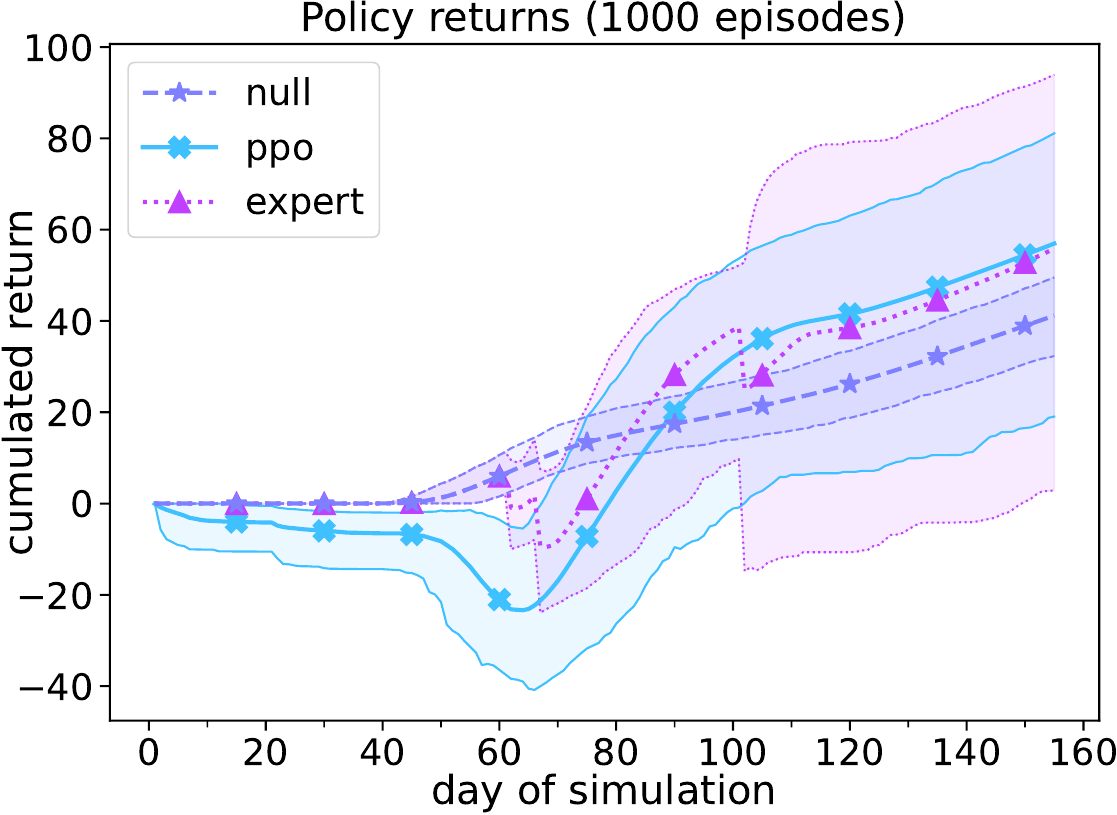}} \label{fig:fertilizationRewards}}%
    \qquad
    \subfloat[\centering 2D histogram of fertilizer applications (the darker the more frequent).]{{\includegraphics[width=.47\textwidth]{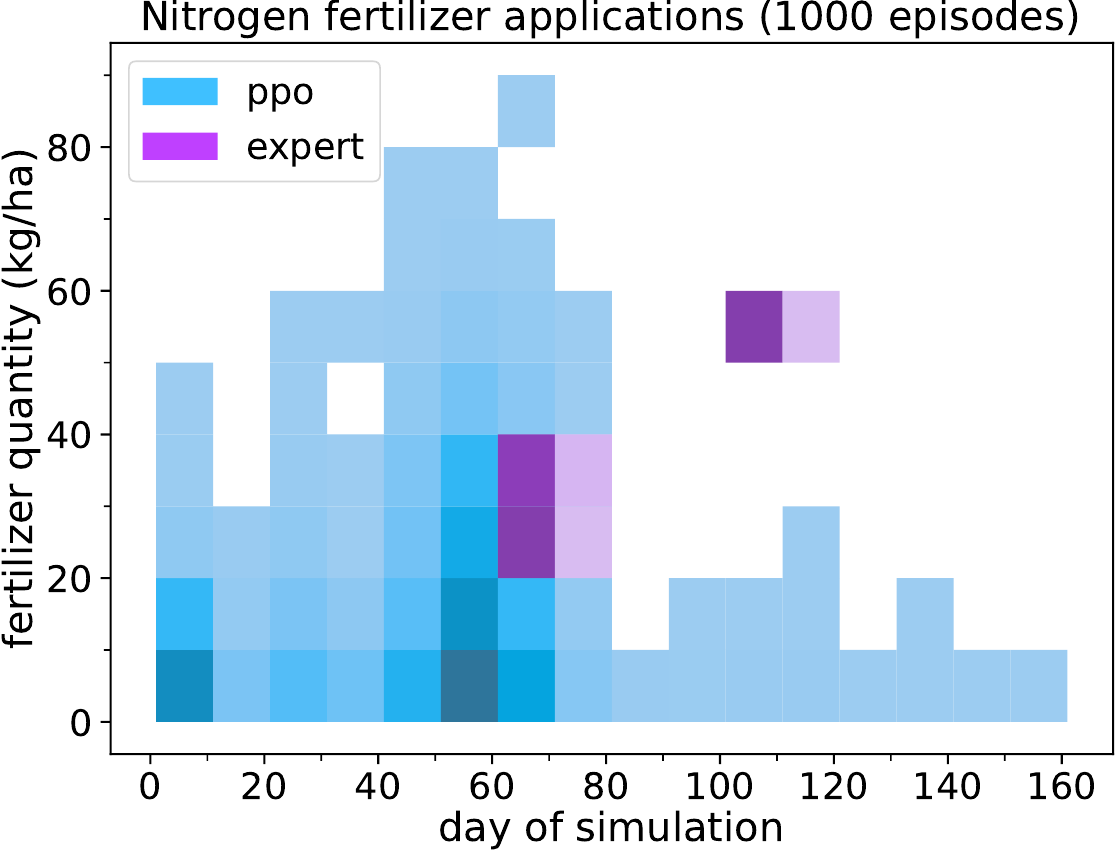}}\label{fig:fertilizationApplications}}%
    \caption{Undiscounted cumulated returns and applications for the fertilization problem.\label{fig:fertilizationExp}}%
\end{figure}

\paragraph{Results}
\figurename~\ref{fig:fertilizationRewards} displays the evolution of undiscounted cumulated rewards (Equation~\ref{eq:J}) of policies, against the day of simulation. PPO ended with the highest mean cumulated return compared to the null and expert policies. PPO cumulated returns were less variable than with the expert policy, as can be seen from the reduced range of values between upper and lower quantiles. \figurename~\ref{fig:fertilizationApplications} provides a 2D histogram of fertilizer applications, against the day of simulation. The darker a cell, the more frequent the fertilizer application. PPO fertilizer applications were more frequent at the beginning of the growing season and around day of simulation 60. The latter application date corresponds to the beginning of the floral initiation stage, see \tablename~\ref{tab:fertilizationStages} in Appendix. Nevertheless, the variability of rates and application dates of PPO indicated that PPO policy did not only depend on days after planting as the expert policy did, but also depended on more factors. Note that while the expert policy was deterministic, the day of simulation of applications showed slight variations. This was because in simulations, the planting date was automatically determined within a time window, depending on soil conditions, depending itself on (stochastic) weather events. Because the expert policy specified fertilizer application dates in days after planting, and not in days of simulation, a shift in planting dates consistently induced a shift in the corresponding day of simulation of fertilizer applications.

\tablename~\ref{tab:fertilizationPerf} shows statistics of the performance indicators detailed in \tablename~\ref{tab:fertilizationTestingFeatures}. As expected, there was no policy that was optimal for all performance criteria. PPO policy exhibited performance trade-offs between the expert and the null policies we deemed satisfying. Grain yield and nitrogen content in grains (a nutritional criteria) were close to the ones of expert policy. On average, PPO policy consumed about 28\% less nitrogen than the expert policy. Consistently, PPO ANE (Equation~\ref{eq:fertilizationEfficiency}) --a key metric of sustainable fertilization-- was about 29\% greater than for the expert policy. From a practical perspective, a good fertilization policy consists in a limited number of applications during an episode, as the expert policy suggests. Indeed, each nitrogen application costs both in terms of fertilizer (as a product of natural gas) and application costs (e.g.\@ mechanized nitrogen broadcasting). The mean number of applications of PPO (about 6) was higher than for the expert policy (3), but still practicable. Finally, PPO policy showed a slighlty lower nitrate leaching than the expert policy, which means less nitrate pollution induced by nitrogen fertilization.

\begin{table}\centering
  \begin{tabular}{lccc}
    \toprule
 & \textbf{null}  & \textbf{expert} & \textbf{PPO}       \\ %\cmidrule(lr){2-4}
grain yield (kg/ha) & 1141.1 (344.0)  & \textbf{3686.5} (1841.0)  & 3463.1 (1628.4)     \\
massic nitrogen in grains (\%) & 1.1 (0.1) & \textbf{1.7} (0.2) & 1.5 (0.3) \\ 
total fertilization (kg/ha) & \textbf{0} (0) & 115.8 (5.2) & 82.8 (15.2)    \\
application number & \textbf{0} (0) & 3.0 (0.1) & 5.7 (1.6)    \\
nitrogen use efficiency (kg/kg) & n.a.  & 22.0 (14.1) & \textbf{28.3} (16.7)    \\
nitrate leaching (kg/ha) & \textbf{15.9} (7.7) & 18.0 (12.0) & 18.3 (11.6)\\
\bottomrule
\end{tabular}
\caption{Mean (st.\@ dev.) fertilization baselines performances computed using 1000 episodes. For each criterion, bold numbers indicate the best performing policy.}
\label{tab:fertilizationPerf}
\end{table}

\paragraph{Discussion}We have shown that with an off-the-shelf \texttt{Stable-Baselines3} PPO implementation, we have been able to learn a relevant fertilization policy that slightly outperforms the expert fertilization policy regarding the objective function. From an agronomic perspective, PPO policy reached superior nitrogen use efficiency, with a substantially reduced nitrogen fertilizer consumption compared to the expert policy, while still yielding maize grain with satisfying quantity and quality. PPO focused nitrogen fertilizer applications at the beginning of the floral initiation stage, where maize nitrogen needs are the greatest and most crucial \citep{hanway1963growth}. The performance of PPO is likely to increase with a proper tuning. Nevertheless:

\begin{itemize}
\item[(1)] The fertilization policy an agent has learned still requires \textit{explainability}. For instance, discovering which are the most important observation variables that determine a fertilizer application, how their values impact fertilization, and if these results are consistent with the agronomic knowledge is a requirement. For crop management decision support systems, user trust is essential \citep{rose2016decision,evans2017data}.
As an example, \citet{garcia1999use} translated an RL agent policy into a set of simple decision rules (e.g.\@ ``if condition 1 or condition 2, then do ...") which were easily interpretable and usable by farmers and/or agronomists.

\item[(2)] In real conditions, each field observation costs.  As an example, the growth stage (\texttt{istage}, \tablename~\ref{tab:fertilizationObservationVariables}), which is an observation variable, would only require a periodic visual inspection of the field. Growth stage is consequently a realistic and inexpensive observation. In contrast, the measure of the daily population nitrogen uptake, necessary to compute the return  (\texttt{trnu}, Equation~\ref{eq:rewardFertilization}), would require destructive plant sampling and extensive laboratory analysis. In case the agent is trained with real field trials, then computing rewards becomes necessary, and might be problematic. Consequently, in latter case, an alternative reward function could be employed. The cost of measuring each observation variable --related to the precision of measurement-- and the frequency of these observations should be minimized for practical applications. 

\item[(3)] Learning a relevant fertilization policy from scratch required {$10^6$} episodes. The stochastic weather time series \texttt{gym-DSSAT} used being sampled from independent and identically distributions, {$10^6$} episodes means {$10^6$} cultivation cycles under different weather conditions. If the objective of the experiment is to design \textit{in silico} fertilization policies, then learning efficiency and field measure costs (2) are not problematic, but remark (1) still applies. If \texttt{gym-DSSAT} is used to mimic real-world conditions and the objective is to design an RL algorithm able to learn/improve from real interactions, then the learning efficiency of the off-the-shelf PPO clearly precludes any straight application in real conditions. Thereby, researchers must reduce the sample complexity of the decision problem, i.e.\@ simplify the problem to reduce the number of samples required to solve this problem, and/or researchers must use/design other RL algorithms with improved learning efficiency \citep[e.g.\@ using demonstration learning][to leverage existing expert policies]{taylor2009transfer}.
\end{itemize}

\subsection{Execution time and reproducibility}

\noindent In this section, we now briefly highlight that \texttt{gym-DSSAT} is a lightweight RL environment and discuss reproducibility issues.

\paragraph{Execution time}We performed all measures of \texttt{gym-DSSAT} execution time for the fertilization task. The mean duration of an episode was $156 \pm 7~\text{days}$ (1 time step was 1 simulated day), averaged over 1000 episodes. We measured the following time executions averaged over 1000 episodes, each episode lasted until 100 time steps. In practice, we insured that all episodes did not end between step 1 and step 100, so environments had to update their state for the 100 time steps. During an episode, actions were randomly sampled from the action space. On a standard 8 core laptop, the mean running time to simulate one day in \texttt{gym-DSSAT} i.e.\@ taking a single step in the environment was $2.56 \pm 0.22 ~\text{ms}$. In comparison, the mean running time of taking a step in \texttt{gym} default \texttt{MuJoCo} (version 2.1.0) environment \texttt{HumanoidStandup-v2} was $0.61 \pm 0.21 ~\text{ms}$. While \texttt{gym-DSSAT} is more expensive in time than typical \texttt{gym} environments, the simulation is still responsive enough for typical usage in RL experiments.

\paragraph{Reproducibility}According to the Association For Computing Machinery (ACM), a computational experiment is said \textit{reproducible} if an ``[\dots] independent group can obtain the same result using the author’s own artifacts''\footnote{Artifact Review and Badging Version 1.1 - August 24, 2020, \url{https://www.acm.org/publications/policies/artifact-review-and-badging-current}}, summarized as ``different team, same experimental setup''.
%We define a study with \textit{reproducible} results as a study which provides all necessary original materials and analysis workflow for other researchers to successfully replicate these results \citep{goodman2016does}. 
Based on our tests, we successfully reproduced the results of the present study on the same platform i.e.\@ on the same hardware and software layers. This means that both results of \texttt{gym-DSSAT} and \texttt{Stable-Baselines3} PPO were reproducible on the same platform. Nevertheless, as a more general reproducibility issue, we cannot guarantee the cross-platform reproducibility. Reaching cross-platform reproducibility is a generally hard issue, even for deterministic software, due to the multiple factors at stake. As an example, compiling \texttt{DSSAT} Fortran code with two different compilers may not result in the same exact \texttt{DSSAT} outputs. This is because the order of multiple arithmetic operations, despite being mathematically commutative, may not follow the same order in practice and the final result might be different because of floating point number rounding effects. To enhance reproducibility, we provide Docker containers for various Linux distributions for \texttt{gym-DSSAT} (see \href{https://rgautron.gitlabpages.inria.fr/gym-dssat-docs/Installation/index.html}{installation instructions}\footnote{\url{https://rgautron.gitlabpages.inria.fr/gym-dssat-docs/Installation/index.html}}).

\section{Concluding remarks}
\label{sec:futureWorks}
In this paper, we briefly presented \texttt{gym-DSSAT}, a Reinforcement Learning (RL) environment for crop management, and exposed uses cases for fertilization and irrigation decision problems. \texttt{gym-DSSAT} is based on \texttt{DSSAT}, a celebrated crop simulator used by worldwide agronomists. To turn the original Fortran \texttt{DSSAT} software into a Python \texttt{gym} environment, we used a recently introduced library, named \texttt{PDI}. Currently, only maize fertilization and irrigation problems are available. \texttt{gym-DSSAT} can be extended to any of the 41 other crops \texttt{DSSAT} currently simulates, such as wheat or sorghum and/or to other crop operations. Further predefined scenarios will be defined to reflect a diversity of soil and climate combinations. Weather forecasts being of major interest for crop management \citep{hoogenboom2000contribution}, short time weather predictions of stochastically generated weather will be provided in the state space.  \texttt{gym-DSSAT} will be connected to \texttt{Ray rllib} \citep{liang2017ray} to enhance environment scalability. 
%We have demonstrated how a standard RL algorithm, PPO, may learn to manage a crop plot.  
For both irrigation and fertilization use cases, we showed that an untuned RL algorithm was able to learn more sustainable practices than the expert policies we considered. 
Beyond the use cases we have provided, further work is still required to tailor RL algorithms to the idiosyncracies of crop management problems. The performance baselines of each decision problem can be iteratively refined, for instance using the expert policy with Transfer Learning \citep{taylor2009transfer} or extra exploration such as with Random Network Distillation \citep{burda2018exploration}. With a limited software development effort, \texttt{PDI} can be used to turn other existing mechanistic models into \texttt{gym} environments, hence opening the doors of a potentially large number of mechanistic models to the RL community. We hope the whole approach we used to create \texttt{gym-DSSAT} will be replicated to other complementary C, C++ or Fortran based crop models, such as STICS \citep{brisson2003overview} and other mechanistic models.

\newpage
\section*{Acknowledgments}
\noindent 
Emilio J.\@ Padrón's work was partially supported through the research projects PID2019-104184RB-I00 funded by MCIN/AEI/10.13039/501100011033, and
ED431C 2021/30 and ED431G 2019/01 funded by Xunta de Galicia. The authors acknowledge the PDI team, in particular Karol Sierocinski, for their help. The authors also acknowledge Bruno Raffin for his support and Essam Morsi for his initial help. Thanks to the \texttt{DSSAT} team, especially Gerrit Hoogenboom and Cheryl Porter for their continuous support. We thank Jacob van Etten for his remarks. We acknowledge the Consultative Group for International Agricultural Research (CGIAR) Platform for Big Data in Agriculture and we especially thank Brian King. Ph.\@ Preux, O-A.\@ Maillard and D.\@ Emukpere acknowledge the support of the Métropole Européenne de Lille (MEL), ANR, Inria, Université de Lille, through the AI chair Apprenf number R-PILOTE-19-004-APPRENF. We acknowledge the AIDA team of the French Agricultural Research Centre for International Development (CIRAD) and the outstanding working environment provided by Inria and the Scool research group.

\bibliography{biblio}

\begin{thebibliography}{}

\bibitem[{\AA}str{\"o}m, 1965]{aastrom1965optimal}
{\AA}str{\"o}m, K.~J. (1965).
\newblock Optimal control of markov processes with incomplete state
  information.
\newblock {\em Journal of Mathematical Analysis and Applications},
  10(1):174--205.

\bibitem[Binas et~al., 2019]{binasreinforcement}
Binas, J., Luginbuehl, L., and Bengio, Y. (2019).
\newblock Reinforcement learning for sustainable agriculture.
\newblock In {\em ICML Workshop Climate Change: How Can AI Help?}

\bibitem[Binder et~al., 2000]{binder2000maize}
Binder, D.~L., Sander, D.~H., and Walters, D.~T. (2000).
\newblock Maize response to time of nitrogen application as affected by level
  of nitrogen deficiency.
\newblock {\em Agronomy Journal}, 92(6):1228--1236.

\bibitem[Brisson et~al., 2003]{brisson2003overview}
Brisson, N., Gary, C., Justes, E., Roche, R., Mary, B., Ripoche, D., Zimmer,
  D., Sierra, J., Bertuzzi, P., Burger, P., et~al. (2003).
\newblock An overview of the crop model stics.
\newblock {\em European Journal of agronomy}, 18(3-4):309--332.

\bibitem[Burda et~al., 2018]{burda2018exploration}
Burda, Y., Edwards, H., Storkey, A., and Klimov, O. (2018).
\newblock Exploration by random network distillation.
\newblock {\em arXiv preprint arXiv:1810.12894}.

\bibitem[Camargo and Kemanian, 2016]{camargo2016six}
Camargo, G.~G. and Kemanian, A.~R. (2016).
\newblock Six crop models differ in their simulation of water uptake.
\newblock {\em Agricultural and forest meteorology}, 220:116--129.

\bibitem[Cassman et~al., 2002]{cassman2002agroecosystems}
Cassman, K.~G., Dobermann, A., and Walters, D.~T. (2002).
\newblock Agroecosystems, nitrogen-use efficiency, and nitrogen management.
\newblock {\em AMBIO: A Journal of the Human Environment}, 31(2):132--140.

\bibitem[Chatelin et~al., 2005]{chatelin2005decible}
Chatelin, M.-H., Aubry, C., Poussin, J.-C., Meynard, J.-M., Mass{\'e}, J.,
  Verjux, N., Gate, P., and Le~Bris, X. (2005).
\newblock D{\'e}cibl{\'e}, a software package for wheat crop management
  simulation.
\newblock {\em Agricultural Systems}, 83(1):77--99.

\bibitem[Chen et~al., 2021]{chen2021reinforcement}
Chen, M., Cui, Y., Wang, X., Xie, H., Liu, F., Luo, T., Zheng, S., and Luo, Y.
  (2021).
\newblock A reinforcement learning approach to irrigation decision-making for
  rice using weather forecasts.
\newblock {\em Agricultural Water Management}, 250:106838.

\bibitem[Dor{\'e} et~al., 2006]{dore2006agronomie}
Dor{\'e}, T., Martin, P., Le~Bail, M., Ney, B., and Roger-Estrade, J. (2006).
\newblock {\em L'agronomie aujourd'hui}.
\newblock Editions Quae.

\bibitem[Dorier et~al., 2016]{dorier2016damaris}
Dorier, M., Antoniu, G., Cappello, F., Snir, M., Sisneros, R., Yildiz, O.,
  Ibrahim, S., Peterka, T., and Orf, L. (2016).
\newblock Damaris: Addressing performance variability in data management for
  post-petascale simulations.
\newblock {\em ACM Transactions on Parallel Computing (TOPC)}, 3(3):1--43.

\bibitem[Duru et~al., 2015]{duru2015implement}
Duru, M., Therond, O., Martin, G., Martin-Clouaire, R., Magne, M.-A., Justes,
  E., Journet, E.-P., Aubertot, J.-N., Savary, S., Bergez, J.-E., et~al.
  (2015).
\newblock How to implement biodiversity-based agriculture to enhance ecosystem
  services: a review.
\newblock {\em Agronomy for sustainable development}, 35(4):1259--1281.

\bibitem[Evans et~al., 2017]{evans2017data}
Evans, K.~J., Terhorst, A., and Kang, B.~H. (2017).
\newblock From data to decisions: helping crop producers build their actionable
  knowledge.
\newblock {\em Critical reviews in plant sciences}, 36(2):71--88.

\bibitem[Freund, 1956]{freund1956introduction}
Freund, R.~J. (1956).
\newblock The introduction of risk into a programming model.
\newblock {\em Econometrica: Journal of the econometric society}, pages
  253--263.

\bibitem[Garcia, 1999]{garcia1999use}
Garcia, F. (1999).
\newblock Use of reinforcement learning and simulation to optimize wheat crop
  technical management.
\newblock In {\em Proceedings of the International Congress on Modelling and
  Simulation (MODSIM’99) Hamilton, New-Zealand}, pages 801--806.

\bibitem[Gautron et~al., 2022]{gautron2022reinforcement}
Gautron, R., Maillard, O.-A., Preux, P., Corbeels, M., and Sabbadin, R. (2022).
\newblock Reinforcement learning for crop management support: Review, prospects
  and challenges.
\newblock {\em Computers and Electronics in Agriculture}, 200:107182.

\bibitem[Godoy et~al., 2020]{godoy2020adios}
Godoy, W.~F., Podhorszki, N., Wang, R., Atkins, C., Eisenhauer, G., Gu, J.,
  Davis, P., Choi, J., Germaschewski, K., Huck, K., et~al. (2020).
\newblock Adios 2: The adaptable input output system. a framework for
  high-performance data management.
\newblock {\em SoftwareX}, 12:100561.

\bibitem[Golemo et~al., 2018]{golemo2018sim}
Golemo, F., Taiga, A.~A., Courville, A., and Oudeyer, P.-Y. (2018).
\newblock Sim-to-real transfer with neural-augmented robot simulation.
\newblock In {\em Conference on Robot Learning}, pages 817--828. PMLR.

\bibitem[Hanway, 1963]{hanway1963growth}
Hanway, J. (1963).
\newblock Growth stages of corn (zea mays, l.) 1.
\newblock {\em Agronomy Journal}, 55(5):487--492.

\bibitem[Hao et~al., 2018]{hao2018seasonal}
Hao, Z., Singh, V.~P., and Xia, Y. (2018).
\newblock Seasonal drought prediction: advances, challenges, and future
  prospects.
\newblock {\em Reviews of Geophysics}, 56(1):108--141.

\bibitem[Hayes et~al., 2021]{hayes2021practical}
Hayes, C.~F., R{\u{a}}dulescu, R., Bargiacchi, E., K{\"a}llstr{\"o}m, J.,
  Macfarlane, M., Reymond, M., Verstraeten, T., Zintgraf, L.~M., Dazeley, R.,
  Heintz, F., et~al. (2021).
\newblock A practical guide to multi-objective reinforcement learning and
  planning.
\newblock {\em arXiv preprint arXiv:2103.09568}.

\bibitem[He et~al., 2012]{he2012identifying}
He, J., Dukes, M.~D., Hochmuth, G.~J., Jones, J.~W., and Graham, W.~D. (2012).
\newblock Identifying irrigation and nitrogen best management practices for
  sweet corn production on sandy soils using ceres-maize model.
\newblock {\em Agricultural Water Management}, 109:61--70.

\bibitem[Hill et~al., 2018]{stable-baselines}
Hill, A., Raffin, A., Ernestus, M., Gleave, A., Kanervisto, A., Traore, R.,
  Dhariwal, P., Hesse, C., Klimov, O., Nichol, A., Plappert, M., Radford, A.,
  Schulman, J., Sidor, S., and Wu, Y. (2018).
\newblock Stable baselines.
\newblock \url{https://github.com/hill-a/stable-baselines}.

\bibitem[Hintjens, 2013]{hintjens2013zeromq}
Hintjens, P. (2013).
\newblock {\em ZeroMQ: messaging for many applications}.
\newblock O'Reilly Media, Inc.

\bibitem[Hoogenboom, 2000]{hoogenboom2000contribution}
Hoogenboom, G. (2000).
\newblock Contribution of agrometeorology to the simulation of crop production
  and its applications.
\newblock {\em Agricultural and forest meteorology}, 103(1-2):137--157.

\bibitem[Hoogenboom et~al., 2019]{hoogenboom2019dssat}
Hoogenboom, G., Porter, C., Boote, K., Shelia, V., Wilkens, P., Singh, U.,
  White, J., Asseng, S., Lizaso, J., Moreno, L., et~al. (2019).
\newblock The dssat crop modeling ecosystem.
\newblock {\em Advances in crop modelling for a sustainable agriculture}, pages
  173--216.

\bibitem[Howell, 2003]{howell2003irrigation}
Howell, T.~A. (2003).
\newblock Irrigation efficiency.
\newblock {\em Encyclopedia of water science}, 467:500.

\bibitem[Hunt and Boote, 1998]{hunt1998data}
Hunt, L.~A. and Boote, K.~J. (1998).
\newblock Data for model operation, calibration, and evaluation.
\newblock In {\em Understanding options for agricultural production}, pages
  9--39. Springer.

\bibitem[Husson et~al., 2021]{Husson2021}
Husson, O., Sarthou, J.-P., Bousset, L., Ratnadass, A., Schmidt, H.-P., Kempf,
  J., Husson, B., Tingry, S., Aubertot, J.-N., Deguine, J.-P., Goebel, F.-R.,
  and Lamichhane, J.~R. (2021).
\newblock Soil and plant health in relation to dynamic sustainment of eh and ph
  homeostasis: A review.
\newblock {\em Plant and Soil}.

\bibitem[Kadam et~al., 2014]{kadam2014agronomic}
Kadam, N.~N., Xiao, G., Melgar, R.~J., Bahuguna, R.~N., Quinones, C.,
  Tamilselvan, A., Prasad, P. V.~V., and Jagadish, K.~S. (2014).
\newblock Agronomic and physiological responses to high temperature, drought,
  and elevated co2 interactions in cereals.
\newblock {\em Advances in agronomy}, 127:111--156.

\bibitem[Kamara et~al., 2003]{kamara2003influence}
Kamara, A., Menkir, A., Badu-Apraku, B., and Ibikunle, O. (2003).
\newblock The influence of drought stress on growth, yield and yield components
  of selected maize genotypes.
\newblock {\em The journal of agricultural science}, 141(1):43--50.

\bibitem[Lapan, 2018]{lapan2018deep}
Lapan, M. (2018).
\newblock {\em Deep Reinforcement Learning Hands-On: Apply modern RL methods,
  with deep Q-networks, value iteration, policy gradients, TRPO, AlphaGo Zero
  and more}.
\newblock Packt Publishing Ltd.

\bibitem[Li et~al., 2019]{li2019excessive}
Li, Y., Guan, K., Schnitkey, G.~D., DeLucia, E., and Peng, B. (2019).
\newblock Excessive rainfall leads to maize yield loss of a comparable
  magnitude to extreme drought in the united states.
\newblock {\em Global change biology}, 25(7):2325--2337.

\bibitem[Liang et~al., 2017]{liang2017ray}
Liang, E., Liaw, R., Nishihara, R., Moritz, P., Fox, R., Gonzalez, J.,
  Goldberg, K., and Stoica, I. (2017).
\newblock Ray rllib: A composable and scalable reinforcement learning library.
\newblock {\em arXiv preprint arXiv:1712.09381}, page~85.

\bibitem[Meisinger and Delgado, 2002]{meisinger2002principles}
Meisinger, J.~J. and Delgado, J.~A. (2002).
\newblock Principles for managing nitrogen leaching.
\newblock {\em Journal of soil and water conservation}, 57(6):485--498.

\bibitem[Meurdesoif et~al., 2013]{meurdesoif2013xios}
Meurdesoif, Y., Ozdoba, H., Caubel, A., and Marti, O. (2013).
\newblock Xios.
\newblock In {\em Second Workshop on Coupling Technologies for Earth System
  Models (CW2013), NCAR, Boulder, CO, USA, available at:
  http://forge.ipsl.jussieu.fr/ioserver/raw-attachment/wiki/WikiStart/XIOS-BOULDER.pdf
  (last access: 5 August 2021)}.

\bibitem[Modak, 2002]{modak2002haber}
Modak, J.~M. (2002).
\newblock Haber process for ammonia synthesis.
\newblock {\em Resonance}, 7(9):69--77.

\bibitem[Morris et~al., 2018]{morris2018strengths}
Morris, T.~F., Murrell, T.~S., Beegle, D.~B., Camberato, J.~J., Ferguson,
  R.~B., Grove, J., Ketterings, Q., Kyveryga, P.~M., Laboski, C.~A., McGrath,
  J.~M., et~al. (2018).
\newblock Strengths and limitations of nitrogen rate recommendations for corn
  and opportunities for improvement.
\newblock {\em Agronomy Journal}, 110(1):1.

\bibitem[Mueller et~al., 2012]{mueller2012closing}
Mueller, N.~D., Gerber, J.~S., Johnston, M., Ray, D.~K., Ramankutty, N., and
  Foley, J.~A. (2012).
\newblock Closing yield gaps through nutrient and water management.
\newblock {\em Nature}, 490(7419):254--257.

\bibitem[NeSmith and Ritchie, 1992]{nesmith1992short}
NeSmith, D. and Ritchie, J. (1992).
\newblock Short-and long-term responses of corn to a pre-anthesis soil water
  deficit.
\newblock {\em Agronomy journal}, 84(1):107--113.

\bibitem[Ng et~al., 1999]{ng1999policy}
Ng, A.~Y., Harada, D., and Russell, S. (1999).
\newblock Policy invariance under reward transformations: Theory and
  application to reward shaping.
\newblock In {\em Icml}, volume~99, pages 278--287.

\bibitem[Overweg et~al., 2021]{overweg2021cropgym}
Overweg, H., Berghuijs, H.~N., and Athanasiadis, I.~N. (2021).
\newblock Cropgym: a reinforcement learning environment for crop management.
\newblock {\em arXiv preprint arXiv:2104.04326}.

\bibitem[Papy, 1998]{papy1998savoir}
Papy, F. (1998).
\newblock Savoir pratique sur les syst{\`e}mes techniques et aide {\`a} la
  d{\'e}cision.
\newblock {\em La conduite du champ cultiv{\'e}. Points de vue d’agronomes.
  IRD}, pages 245--259.

\bibitem[Puterman, 1994]{puterman1994markov}
Puterman, M.~L. (1994).
\newblock {\em Markov decision processes: discrete stochastic dynamic
  programming}.
\newblock John Wiley \& Sons.

\bibitem[Randl{\o}v and Alstr{\o}m, 1998]{randlov1998learning}
Randl{\o}v, J. and Alstr{\o}m, P. (1998).
\newblock Learning to drive a bicycle using reinforcement learning and shaping.
\newblock In {\em ICML}, volume~98, pages 463--471. Citeseer.

\bibitem[Richardson, 1985]{richardson1985weather}
Richardson, C. (1985).
\newblock Weather simulation for crop management models.
\newblock {\em Transactions of the ASAE}, 28(5):1602--1606.

\bibitem[Rose et~al., 2016]{rose2016decision}
Rose, D.~C., Sutherland, W.~J., Parker, C., Lobley, M., Winter, M., Morris, C.,
  Twining, S., Ffoulkes, C., Amano, T., and Dicks, L.~V. (2016).
\newblock Decision support tools for agriculture: Towards effective design and
  delivery.
\newblock {\em Agricultural systems}, 149:165--174.

\bibitem[Roussel et~al., 2017]{roussel2017pdi}
Roussel, C., Keller, K., Gaalich, M., Gomez, L.~B., and Bigot, J. (2017).
\newblock {PDI}, an approach to decouple {I/O} concerns from high-performance
  simulation codes.
\newblock Working paper.

\bibitem[Schulman et~al., 2017]{schulman2017proximal}
Schulman, J., Wolski, F., Dhariwal, P., Radford, A., and Klimov, O. (2017).
\newblock Proximal policy optimization algorithms.
\newblock {\em arXiv preprint arXiv:1707.06347}.

\bibitem[Sebillotte, 1974]{sebillotte1974agronomie}
Sebillotte, M. (1974).
\newblock Agronomie et agriculture. essai d’analyse des t{\^a}ches de
  l’agronome.
\newblock {\em Cahiers Orstom, s{\'e}rie biologie}, 24:3--25.

\bibitem[Sebillotte, 1978]{sebillotte1978itineraires}
Sebillotte, M. (1978).
\newblock Itin{\'e}raires techniques et {\'e}volution de la pens{\'e}e
  agronomique.
\newblock {\em CR Acad. Agric. Fr}, 64(11):906--914.

\bibitem[Shibu et~al., 2010]{shibu2010lintul3}
Shibu, M.~E., Leffelaar, P.~A., Van~Keulen, H., and Aggarwal, P.~K. (2010).
\newblock Lintul3, a simulation model for nitrogen-limited situations:
  Application to rice.
\newblock {\em European Journal of Agronomy}, 32(4):255--271.

\bibitem[Sokolowski and Banks, 2012]{sokolowski2012handbook}
Sokolowski, J.~A. and Banks, C.~M. (2012).
\newblock {\em Handbook of real-world applications in modeling and simulation},
  volume~2.
\newblock John Wiley \& Sons.

\bibitem[Soltani and Hoogenboom, 2003]{soltani2003statistical}
Soltani, A. and Hoogenboom, G. (2003).
\newblock A statistical comparison of the stochastic weather generators wgen
  and simmeteo.
\newblock {\em Climate Research}, 24(3):215--230.

\bibitem[Spaan, 2012]{spaan2012partially}
Spaan, M.~T. (2012).
\newblock Partially observable markov decision processes.
\newblock In {\em Reinforcement Learning}, pages 387--414. Springer.

\bibitem[Stone, 1974]{stone1974cross}
Stone, M. (1974).
\newblock Cross-validatory choice and assessment of statistical predictions.
\newblock {\em Journal of the Royal Statistical Society: Series B
  (Methodological)}, 36(2):111--133.

\bibitem[Sun et~al., 2017]{sun2017reinforcement}
Sun, L., Yang, Y., Hu, J., Porter, D., Marek, T., and Hillyer, C. (2017).
\newblock Reinforcement learning control for water-efficient agricultural
  irrigation.
\newblock In {\em 2017 IEEE International Symposium on Parallel and Distributed
  Processing with Applications and 2017 IEEE International Conference on
  Ubiquitous Computing and Communications (ISPA/IUCC)}, pages 1334--1341. IEEE.

\bibitem[Sutton and Barto, 2018]{sutton2018reinforcement}
Sutton, R.~S. and Barto, A.~G. (2018).
\newblock {\em Reinforcement learning: An introduction}.
\newblock MIT press.

\bibitem[Taylor and Stone, 2009]{taylor2009transfer}
Taylor, M.~E. and Stone, P. (2009).
\newblock Transfer learning for reinforcement learning domains: A survey.
\newblock {\em Journal of Machine Learning Research}, 10(7).

\bibitem[Tintner, 1955]{tintner1955stochastic}
Tintner, G. (1955).
\newblock Stochastic linear programming with applications to agricultural
  economics.
\newblock In {\em Proceedings of the Second Symposium in Linear Programming},
  volume~1, pages 197--228. National Bureau of Standards Washington, DC.

\bibitem[Vanlauwe et~al., 2011]{vanlauwe2011agronomic}
Vanlauwe, B., Kihara, J., Chivenge, P., Pypers, P., Coe, R., and Six, J.
  (2011).
\newblock Agronomic use efficiency of n fertilizer in maize-based systems in
  sub-saharan africa within the context of integrated soil fertility
  management.
\newblock {\em Plant and soil}, 339(1):35--50.

\bibitem[Wallach et~al., 2018]{wallach2018working}
Wallach, D., Makowski, D., Jones, J.~W., and Brun, F. (2018).
\newblock {\em Working with dynamic crop models: methods, tools and examples
  for agriculture and environment}.
\newblock Academic Press.

\bibitem[Wang et~al., 2020]{wang2020deep}
Wang, L., He, X., and Luo, D. (2020).
\newblock Deep reinforcement learning for greenhouse climate control.
\newblock In {\em 2020 IEEE International Conference on Knowledge Graph
  (ICKG)}, pages 474--480. IEEE.

\end{thebibliography}

\appendix
\section{Irrigation use case}
\label{sec:irrigationUseCase}
\noindent We provide a simple baseline for the irrigation problem, as introduced in Section~\ref{subsec:defaultTasks}.

\paragraph{Methods}Overall, the irrigation use case follows the same methods than the fertilization use case (Section~\ref{subsec:fertilizationUseCase}). It only differs from the fertilization use case in the observation space and return function. \tablename~\ref{tab:irrigationObservationVariables} details the default observation space for the irrigation problem. Denoting $\texttt{topwt}(t,t+1)$ the above the ground population biomass change between $t$ and $t\!+\!1$ (kg/ha); and $\texttt{amir}(t)$ the irrigated water on day $t$ (L/\SI{}{\metre\squared}), the default irrigation return function was defined as:
\begin{equation}
r(t) = \underbrace{\texttt{topwt}(t, t + 1)}_{\substack{\text{change in above} \\ \text{the ground biomass}}} - \underbrace{15}_{\substack{\text{penalty} \\ \text{factor}}} \times ~\underbrace{\texttt{amir}(t)}_{\substack{\text{irrigated water} \\ \text{quantity}}}
\label{eq:rewardIrrigation}
\end{equation}
We considered 3 different policies:
\begin{itemize}
\item The `null' policy that never irrigated, which corresponded to rainfed crops. Agronomists may measure the effect of an irrigation policy as a performance gain compared to the null policy, in order to decouple the effect of irrigation from the effect of rainfall \citep{howell2003irrigation}.
\item The second baseline was the ``expert" policy, which was an approximation of the irrigation policy of the original maize field experiment \citep[][\texttt{UFGA8201} experiment number \#3]{hunt1998data}, see Section~\ref{subsec:defaultTasks}. As \tablename~\ref{tab:irrigationExpertPolicy} shows, this policy consisted in sixteen deterministic water applications, which only depended on the number of days after planting. In contrast with the fertilization expert policy (\tablename~\ref{tab:fertilizationExpertPolicy}), this irrigation expert policy was a simplistic approximation of the true expert policy of the original field experiment. Indeed, the true expert policy, unavailable, was likely to depend on more factors (e.g. soil moisture, or days without effective rainfall in a given growth stage) rather than only on days after planting. Nevertheless, the irrigation policy in \tablename~\ref{tab:irrigationExpertPolicy} was still a convenient baseline for this experiment.

\begin{table}\centering
  \begin{tabular}{cc}
    \toprule
    \textbf{DAP} & \textbf{quantity (L/\SI{}{\metre\squared})} \\
    6 & 13 \\
    20 & 10 \\
    37 & 10 \\
    50 & 13 \\
    54 & 18 \\
    65 & 25 \\
    69 & 25 \\
    72 & 13 \\
    75 & 15 \\
    77 & 19 \\
    80 & 20 \\
    84 & 20 \\
    91 & 15 \\
    101 & 19 \\
    104 & 4 \\    
    105 & 25 \\
\bottomrule
\end{tabular}
\caption{Expert irrigation policy. `DAP' stands for Day After Planting.}
\label{tab:irrigationExpertPolicy}
\end{table}
\item The policy learned by PPO.
%We trained PPO using exactly the same procedure than described in Section~\ref{subsec:fertilizationUseCase}.
\end{itemize}
For an irrigation policy $\pi$, denoting $\texttt{grnwt}^{\pi}$ the dry matter grain yield of the policy $\pi$ (kg/ha) ; $\texttt{grnwt}^{0}$ the dry matter grain yield with no fertilization (kg/ha); and $\texttt{totir}^{\pi}$ the total irrigated water with policy $\pi$ (L/\SI{}{\metre\squared}), we define the 
water use efficiency \citep[WUE, Equation 15,][]{howell2003irrigation} as:
\begin{equation}
        \text{WUE}^{\pi}(t)=10 \times \frac{\texttt{grnwt}^{\pi}(t) - \texttt{grnwt}^0(t)}{\texttt{totir}^{\pi}(t)}
        \label{eq:irrigationEfficiency}
\end{equation}
Similarly to the fertilization use case, \tablename~\ref{tab:irrigationTestingFeatures} shows the performance indicators we considered for the irrigation problem. In particular, for excessive irrigation, nitrate leaching may increase \citep{meisinger2002principles}. Thus, nitrate leaching is a pollution performance indicator of irrigation.

\begin{table}\centering
  \begin{tabular}{cll}
    \toprule
 \textbf{variable}  & \textbf{definition} & \textbf{comment}       \\ %\cmidrule(lr){2-4}
\texttt{grnwt} & grain yield (kg/ha) & quantitative objective to be maximized \\
\texttt{totir} & total irrigation (L/\SI{}{\metre\squared}) & cost to be minimized \\
-- & application number & cost to be minimized \\
-- & water use efficiency (kg/\SI{}{\cubic\meter}) & agronomic criteria to be maximized \\
\texttt{runoff} & running-off water (L/\SI{}{\metre\squared}) & loss to be minimized \\
\texttt{cleach} & nitrate leaching (kg/ha) & loss/pollution to be minimized \\
\bottomrule
\end{tabular}
\caption{Performance indicators for irrigation policies. An hyphen means \texttt{gym-DSSAT} does not directly provide the variable, but it can be easily derived.}
\label{tab:irrigationTestingFeatures}
\end{table}

\begin{table}\centering
  \begin{tabular}{cl}
    \toprule
    & \textbf{definition} \\
    \texttt{istage} & \texttt{DSSAT} maize growing stage \\
    \texttt{vstage} & vegetative growth stage (number of leaves) \\
    \texttt{grnwt} & grain weight dry matter (kg/ha) \\
    \texttt{topwt} & above the ground population biomass (kg/ha) \\
    \texttt{xlai} & plant population leaf area index (\SI{}{\metre\squared} leaf/\SI{}{\metre\squared} soil) \\
    \texttt{tmax} & maximum temperature for current day \SI{}{\celsius} \\
    \texttt{srad} & solar radiation during the current day (MJ/\SI{}{\metre\squared}/day) \\
    \texttt{dtt} & growing degree days for current day (\SI{}{\celsius}/day) \\
    \texttt{dap} & duration after planting (day) \\
    \texttt{sw} & volumetric soil water content in soil layers (\SI{}{\cubic\cm} [water] / \SI{}{\cubic\cm} [soil]) \\
    \texttt{ep} & actual plant transpiration rate (L/\SI{}{\metre\squared}/day) \\
    \texttt{wtdep} & depth to water table (cm) \\
    \texttt{rtdep} & root depth (cm) \\
    \texttt{totir} & total irrigated water (L/\SI{}{\metre\squared}) \\
\bottomrule
\end{tabular}
\caption{Default observation space for the irrigation task.}
\label{tab:irrigationObservationVariables}
\end{table}

\begin{figure}%
    \centering
    \subfloat[\centering Mean cumulated return of each of the 3 policies against the day of the simulation. Shaded area displays the $\lbrack{}0.05, 0.95\rbrack$ quantile range for each policy.]{{\includegraphics[width=.47\textwidth]{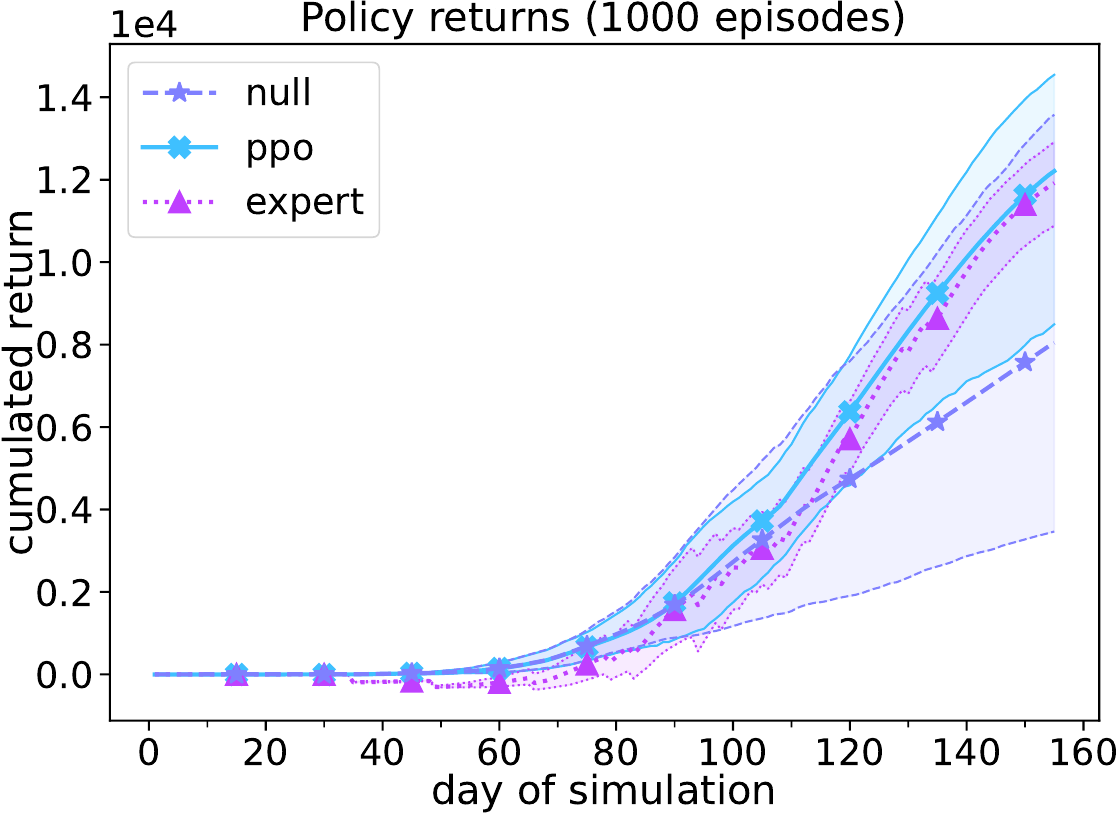}} \label{fig:irrigationRewards}}%
    \qquad
    \subfloat[\centering 2D histogram of irrigations (the darker the more frequent).]{{\includegraphics[width=.47\textwidth]{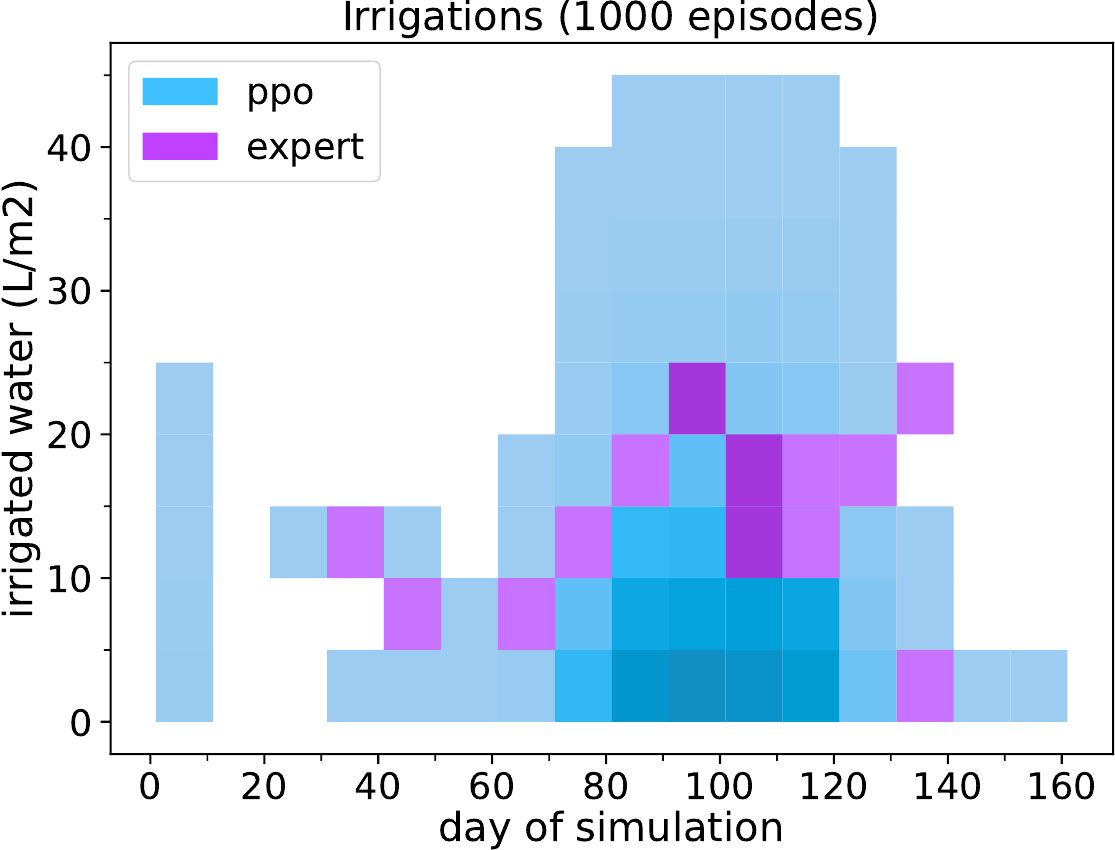}}\label{fig:irrigationApplications}}%
    \caption{Undiscounted cumulated returns and applications for the irrigation problem.\label{fig:irrigationExp}}%
\end{figure}

\paragraph{Results}Regarding the maximization of the undiscounted objective function, PPO showed the best mean performance and slightly outperformed the expert policy, but had an increase variance than the latter, see the wider range of values between upper and lower quantiles in \figurename~\ref{fig:irrigationRewards}. PPO water applications were more frequently found between days 80 and 120 of the simulation, which mostly corresponds to the grain filling stage, see \tablename~\ref{tab:irrigationStages} in Appendix. During this period, in most cases, PPO irrigated less water than the expert policy, see \figurename~\ref{fig:irrigationApplications}.
As indicated in \tablename~\ref{tab:irrigationPerf}, PPO irrigation policy consumed in average about 49\% less water than the expert policy, while maintaining a maize grain yield close to the one of the expert policy. Consistently, the water use efficiency (Equation~\ref{eq:irrigationEfficiency}) of PPO policy was 54\% higher than for the expert policy. Total nitrate leaching for PPO policy was very close to the null policy, and was about 24\% less than for the expert policy. The number of water applications were similar for both expert and PPO policies. Null, expert, and PPO policies had similar water runoff, indicating no water loss due to excessive irrigation of expert or PPO policies.

\begin{table}\centering
  \begin{tabular}{lccc}
    \toprule
 & \textbf{null}  & \textbf{expert} & \textbf{PPO}       \\ %\cmidrule(lr){2-4}
grain yield (kg/ha) & 3734.8 (1852.2)  & \textbf{8306.6} (562.0)  & 7082.2 (1455.7)     \\
total irrigation (\SI{}{\litre}/\SI{}{\metre\squared}) & \textbf{0} (0) & 264.0 (0) & 133.8 (40.3)    \\
application number & \textbf{0} (0) & 16.0 (0.0) & 16.2 (3.7)    \\
water use efficiency (kg/\SI{}{\cubic\meter}) & n.a.  & 17.3 (7.1) & \textbf{26.3} (13.6)    \\
runoff (L/\SI{}{\metre\squared}) & \textbf{0.4} (3.5) & \textbf{0.4} (3.5) & \textbf{0.4} (3.5)\\
nitrate leaching (kg/ha) & \textbf{18.5} (12.6) & 24.6 (9.0) & 18.7 (9.6)\\
\bottomrule
\end{tabular}
\caption{Mean (st.\@ dev.) irrigation baselines performances computed using 1000 episodes. For each criterion, bold numbers indicate the best performing policy.}
\label{tab:irrigationPerf}
\end{table}

\paragraph{Discussion}PPO showed a great efficiency advantage over the expert policy, while maintaining a comparable average grain yield. Water applications of PPO were most frequently focused during maize anthesis period, where maize water needs are the greatest and most crucial with respect to grain yield \citep{nesmith1992short}. Because the expert irrigation policy was likely to be a poor simplification of the real expert irrigation strategy, the advantage PPO irrigation strategy showed might be overly optimistic. However, because PPO has shown largely reduced irrigated water and nitrate leaching, we still deem these results interesting. An alternative baseline could be to reproduce the built-in automatic irrigation policy implemented in \texttt{DSSAT} \citep{hoogenboom2019dssat}, and compare its performance to the irrigation policy of PPO.

\begin{table}[t]\centering
  \begin{tabular}{clccc}
    \toprule
   \texttt{istage} & meaning & \textbf{null} & \textbf{expert} & \textbf{ppo} \\
   8 & 50\% of plants germinated & 28 (0) & 28 (0) & 28 (0) \\
   9 & 50\% of plants with some part visible at soil surface & 29 (0) & 29 (0) & 29 (0) \\
   1 & end of juvenile stage & 40 (3) & 40 (3) & 40 (3) \\
   2 & 50\% of plants completed floral initiation  & 64 (4) & 64 (4) & 64 (4) \\
   3 & 50\% of plants with some silks visible outside husks & 69 (4) & 69 (4) & 69 (4) \\
   4 & beginning of grain filling & 110 (4) & 110 (4) & 110 (4) \\
   5 & end of grain filling & 120 (4) & 120 (4) & 120 (4) \\
   6 & 50\% of plants at harvest maturity  & 158 (4) & 158 (4) & 158 (4) \\
\bottomrule
\end{tabular}
\caption{Mean (st.\@ dev.) days of simulation to reach growth stages for the irrigation problem (1000 episodes).}
\label{tab:irrigationStages}
\end{table}

\cleardoublepage
\section{Fertilization use case complement}
\label{sec:fertilizationUseCaseComplement}
\noindent Table~\ref{tab:fertilizationStages} provides statistics about the growth stages for the three policies of the fertilization use case.

\begin{table}[t]\centering
  \begin{tabular}{clccc}
    \toprule
   \texttt{istage} & meaning & \textbf{null} & \textbf{expert} & \textbf{ppo} \\
   8 & 50\% of plants germinated & 22 (1) & 22 (1) & 22 (1) \\
   9 & 50\% of plants with some part visible at soil surface & 23 (1) & 23 (1) & 23 (1) \\
   1 & end of juvenile stage & 34 (3) & 34 (3) & 34 (3) \\
   2 & 50\% of plants completed floral initiation  & 60 (5) & 60 (5) & 60 (5) \\
   3 & 50\% of plants with some silks visible outside husks & 65 (5) & 65 (5) & 65 (5) \\
   4 & beginning of grain filling & 107 (4) & 107 (4) & 107 (4) \\
   5 & end of grain filling & 117 (4) & 117 (4) & 117 (4) \\
   6 & 50\% of plants at harvest maturity  & 155 (5) & 155 (5) & 155 (5) \\
\bottomrule
\end{tabular}
\caption{Mean (st.\@ dev.) days of simulation to reach growth stages for the fertilization problem (1000 episodes).}
\label{tab:fertilizationStages}
\end{table}

\vfill
\end{document}